\documentclass[man,floatsintext]{apa7}
\usepackage[utf8]{inputenc}
\usepackage[export]{adjustbox}

\usepackage{amsmath,amssymb}
\usepackage{mathtools}
\usepackage{bm}

\usepackage{xcolor}
\usepackage{color}

\usepackage{float}
\usepackage{graphicx}
\usepackage{caption}
\usepackage{subcaption}

\usepackage[inline,shortlabels]{enumitem}

\usepackage{array}
\usepackage{booktabs}
\usepackage{multicol}
\usepackage{multirow}
\usepackage{setspace}
\usepackage{afterpage}
\usepackage{longtable}
\usepackage{tabu}
\usepackage{makecell}

\usepackage[english]{babel}
\usepackage{csquotes}
\usepackage[backend=biber,uniquename=false,style=apa,isbn=false,url=false,eprint=false]{biblatex}
\DefineBibliographyStrings{english}{%
  bibliography = {References},
}

\usepackage{hyperref}
\usepackage{epigraph}
\usepackage{ragged2e}

\title{The End of Radical Concept Nativism}
\shorttitle{End of Radical Concept Nativism}
\authorsnames{Joshua S. Rule,Steven T. Piantadosi}
\affiliation{{Department of Psychology, University of California, Berkeley}}
\date{}%
\authornote{%
 We thank Samuel Cheyette, Benjamin Pitt, Emily Sanford, and Elizabeth Spelke for helpful comments and discussion. Photo in Figure~\ref{fig:fierljeppen} courtesy of Wikimedia Commons. This work was supported by the National Science Foundation (STP, grant number 2201843). The authors declare no competing interests.%

Correspondence concerning this article should be addressed to Joshua S. Rule, Department of Psychology, University of California, Berkeley, 2121 Berkeley Way, Berkeley, CA 94704, USA\@. Email: rule@berkeley.edu.%
}
\abstract{%
  Though humans seem to be remarkable learners, arguments in cognitive science and philosophy of mind have long maintained that learning something fundamentally new is impossible. Specifically, Jerry Fodor's arguments for radical concept nativism hold that most, if not all, concepts are innate and that what many \emph{call} concept learning never actually leads to the acquisition of new concepts. These arguments have deeply affected cognitive science, and many believe that the counterarguments to radical concept nativism have been either unsuccessful or only apply to a narrow class of concepts. This paper first reviews the features and limitations of prior arguments. We then identify three critical points---related to issues of expressive power, conceptual structure, and concept possession---at which the arguments in favor of radical concept nativism diverge from describing actual human cognition. We use ideas from computer science and information theory to formalize the relevant ideas in ways that are arguably more scientifically productive. We conclude that, as a result, there is an important sense in which people do indeed learn new concepts.%
}

\begin{document}

\maketitle

\epigraph{The very \emph{idea} of concept learning is, I think, confused.}{---Jerry Fodor (\citeyear{fodor2008lot})}

\section{Introduction}

How people come to possess their immense conceptual repertoire is a central question in cognitive science. Many people have concepts supporting intuitive theories of biological and physical phenomena, concepts for formal inference in mathematics and science, a lexicon with tens of thousands of items, and even concepts supporting complex perceptual, motor, and social skills. Some concepts are laden with nuance and apply only in particular circumstances (e.g.\ \textsc{saunter}, \textsc{schadenfreude}); others are so versatile as to be difficult to describe comprehensively (e.g.\ \textsc{stuff}, \textsc{action}). New concepts come into use to keep pace with social change (e.g.\ \textsc{xeriscaping}) and technological innovation (e.g.\ \textsc{blockchain}), while others fade away with disuse (e.g.\ \textsc{ether}, \textsc{phlogiston}). Some concepts refer only to abstract entities (e.g.\ \textsc{justice}, \textsc{set}), compositional operations in language (e.g.\ \textsc{for}, \textsc{who}), or to imaginary entities entirely lacking physical referents (e.g.\ \textsc{perpetual motion}, \textsc{cyclops}).

Among the basic empirical facts bearing on the question of where concepts come from is that adults \emph{seem} to have many more concepts than children. Children, for example, require many years to master color terms \parencite{sandhofer1999learning,wagner2013slow}, the various natural kinds like \textsc{dalmatian}, \textsc{plant}, and \textsc{fork} \parencite{gelman1991language}, and mathematical concepts like natural number \parencite{carey2009origins,davidson2012does,carey2019ontogenetic,o2021cultural}. By contrast, adults generally possess these and a wide range of other concepts. Often, these concepts support specialized uses: this can be seen in both cross-cultural variation in words \parencite{regier2016languages,kemp2018semantic,lim2024computational} and lexical items shared only within specific speech communities, i.e.\ jargon \parencite{clark1998communal}.

\begin{figure}[t]
  \centering
  \includegraphics[width=0.66\textwidth]{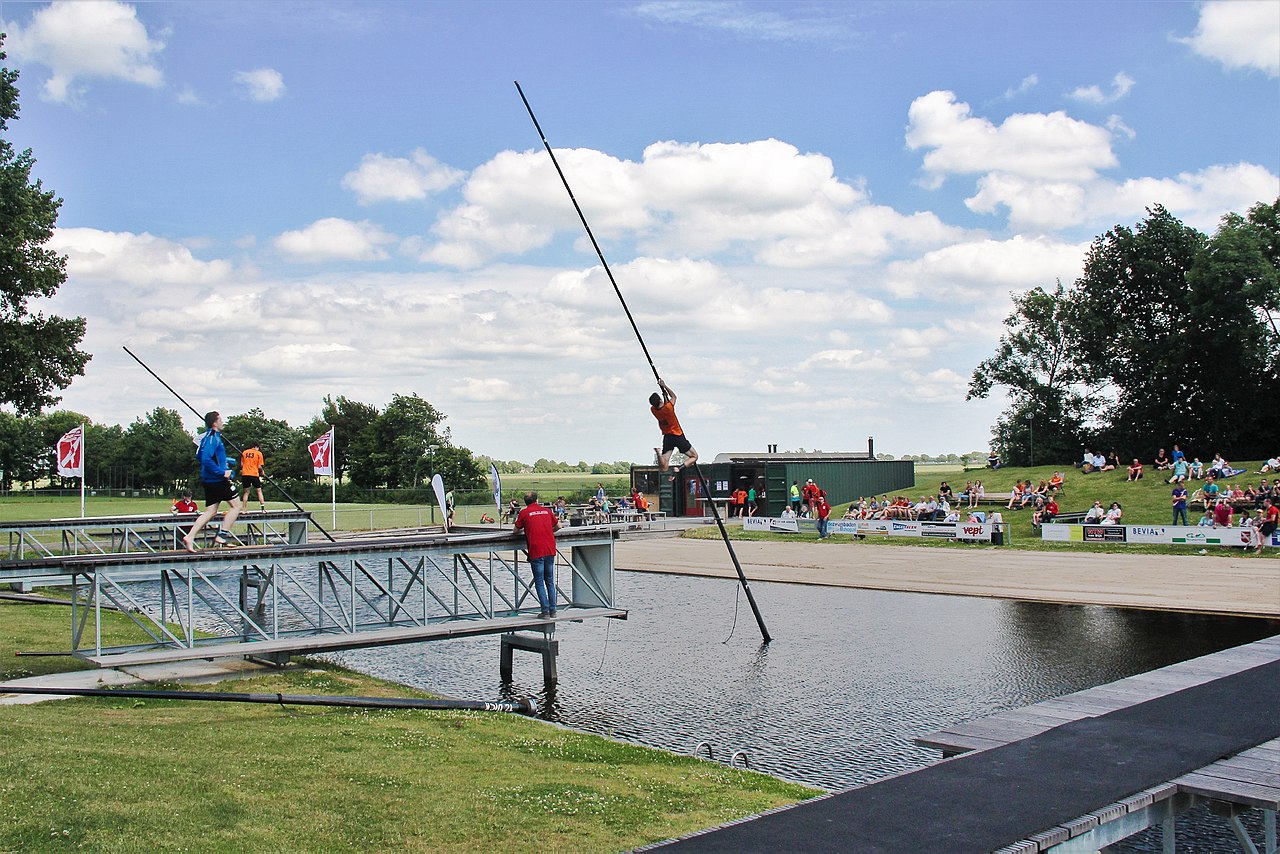}
  \caption{\textsc{Fierljeppen} is a sport where contestants vault a canal by jumping to a pivoting pole and safely dismounting on the other side. The goal is to travel the greatest horizontal distance possible, typically by climbing the pole as it pivots.
  }\label{fig:fierljeppen}
\end{figure}

By far the dominant intuition about the origin of these concepts is that they are acquired through learning. Many people distinctly remember learning things like what an \textsc{integral} or a \textsc{mortgage} is. We can initiate this sort of learning right now in most readers. All we have to do is explain that there is a Dutch sport called \textsc{fierljeppen} where players compete to vault themselves as far as possible across a canal by jumping to a pivoting pole and then safely dismounting on the other side (Figure~\ref{fig:fierljeppen}) \parencite{heck2020fierljeppen}. Readers who had not heard of this sport now have at least a rudimentary understanding and representation of it.

These experiences notwithstanding, some of the most influential arguments in the history of cognitive science hold that all such learning is \emph{impossible}. More specifically, Jerry Fodor (\citeyear{fodor1975language,fodor1980impossibility,fodor1981present,fodor1998concepts}) argued that nearly all lexical concepts---those captured with a single word or morpheme---are innate. These include concepts associated with sensory processing or core cognition, such as \textsc{red}, \textsc{smooth}, \textsc{object}, and \textsc{agent}. More surprisingly, they also include things like \textsc{hippie}, \textsc{floss}, and---famously---\textsc{carburetor}. Even more surprisingly, they include nearly all of the more than 1.4 million entries in the English Wiktionary \parencite{wiktionary2023}, as well as most \emph{possible} words, including those (like \textsc{fierljeppen}) from other languages. According to Fodor, these concepts are innate, as is the full set of concepts capturing technical and cultural innovations still to come. The only concepts which these arguments perhaps leave to be learned are those which are quite clearly constructed of simpler parts, such as \textsc{purple accordion} (\textsc{purple} + \textsc{accordion}) or \textsc{unthinkable} (\textsc{un-} + \textsc{think} + \textsc{-able})\@. \textcite{fodor2008lot} later argues, however, that even these structured concepts cannot be learned. This view that all concept learning is impossible is known as \emph{radical concept nativism}.

Here, we begin by describing three of Fodor's arguments for radical concept nativism. We then review previous responses, noting that while some are more successful than others, none has been generally accepted as refuting the conclusion that nearly all lexical concepts are innate. The heart of the paper is a new argument against radical concept nativism. It begins by identifying three ways in which the arguments for radical concept nativism fail to capture the situation in which human cognitive development actually takes place. First, we argue that it is unlikely that humans have expressive power left to gain, challenging the assumption that concept learning should or could increase theoretical expressive power. At the same time, we note that, in practice, human cognition faces strong resource limitations, and that it is almost certain that compositional learning leads people to experience meaningful changes in expressive power. Second, we challenge the assumption that most concepts are atomic by demonstrating that concepts must be describable in terms that assign them internal computational structure. We further argue that it is very likely that they do in fact have such structure, perhaps being composed of subconceptual parts. Third, we challenge binary notions of concept possession by arguing that they fail to capture the information dynamics at play in human cognition. We instead propose a concrete, mathematical way to think about concept possession which draws on ideas in computer science and information theory. We then show how these three arguments suggest a revised vision of human cognitive development. We conclude that the arguments for radical concept nativism perhaps pose intriguing philosophical puzzles but do not apply to computationally-sophisticated systems like those which have been proposed for inductive inference \parencite{solomonoff1964formal}, artificial intelligence \parencite{hutter2004universal}, and cognitive science \parencite{turing1936computable,turing1950computing}. In these kinds of systems, concept learning is commonplace and in fact necessary.

\section{Fodor's arguments for radical concept nativism} \label{sec:fodor}

Fodor wrote extensively on learning and nativism and his views evolved over time. Moreover, the structure and emphasis in his arguments also changed as needed to fit the demands of the broader discussions in which they appeared. We are not historians of science, and it is not our intention to trace the detailed trajectory of Fodor's views here. Instead, we consider three versions that help to clarify what is at stake in claims about radical concept nativism specifically. The first comes from \textcite{fodor1975language} and presents less an argument for radical concept nativism than an argument against increases in the expressive power of minds' internal representational systems. The second comes from \textcite{fodor1981present} and is perhaps the version of the argument which has had the most impact on cognitive science. The third comes from \textcite{fodor2008lot} and takes an extreme stance in an attempt to both preserve the argument against learning while making some form of concept acquisition possible.

\subsection{Version 1: True Learning and Innate Expressive Power}

\textcite{fodor1975language} says little about radical concept nativism per se. His central argument is instead that psychology cannot be meaningfully reduced either to behaviorism or to neuroscience, because language-like systems of internal representation are required to explain even basic psychological and behavioral phenomena. In this context, however, Fodor asks what this \emph{mentalese} or \emph{language of thought} might be like, and in particular what learning might be like for a mind using a language of thought. While he argues that the answer to this question is largely empirical, Fodor's discussion makes two key contributions to the discussion of radical concept nativism.

The first major contribution Fodor makes here is to identify \emph{concept learning} by contrasting it with other ways in which individuals might come to possess concepts. To begin, some concepts might be \emph{innate}. While never explicitly defining what it means to be innate (reinforcing the point that radical concept nativism is simply \emph{not} a central concern here), this work consistently contrasts cases of innateness with cases of learning, favoring the idea that, ``the language of thought is known{\ldots}but not learned. That is, it is innate,'' and that, ``the internal code is not learned but innately given.\@'' A good analogy might be the sense in which the primitive operations and laws of composition are ``innately'' built into programming languages \parencite{rule2020child}.

It might appear that Fodor is proposing a simple binary alternative, in which concepts are either learned or innate, but he explicitly denies this claim. He instead argues that any concept which is not innate must come to be possessed as the result of some kind of \emph{concept acquisition}. Fodor believed that acquisition might sometimes be possible \emph{without} learning, a phenomenon he saw in Piagetian accounts of development and in changing cognitive constraints (e.g.\ increases in working memory or processing speed). Fodor also considers hypothetical cases of acquiring concepts without learning, such as being hit on the head, taking certain drugs, or surgically manipulating the brain. While these are unlikely to play a major role empirically, the broader point is that we can imagine scenarios where there is no principled relationship between stimulus and the acquired concept. For Fodor, true learning requires exactly this sort of principled connection, where the data somehow confirms the content of what is learned in a way that can be explained using psychological vocabulary. If the explanation is just a brute fact of physics or neurology, then it is not learning for Fodor.\footnote{The concern here is to avoid making arbitrary data-to-content relations---like neurosurgery, bumps on the head, or acquiring \textsc{saturn} the first time you see a swan---into learning mechanisms. Learning must instead be the result of a rational and psychological process.}

The only computational process Fodor sees that could make the right sorts of inductive inferences is to form and test hypotheses. The classic concept learning experiments of  \textcite{bruner1956study}, \textcite{shepard1961learning}, \textcite{nosofsky1994comparing}, and \textcite{feldman2000minimization} are good examples to have in mind. In these, participants might see examples of objects in a low-dimensional feature space (e.g.\ shape, color, size) and learn to label them according to some unobserved logical concept (e.g.\ a ``dax'' is something that is \textsc{small or red}). In this example, learners search a hypothesis space defined by a few primitive\footnote{The word ``primitive'' gets used in many ways in discussions of mental content. We use the word here to pick out objects which have no internal structure comprised of the same kinds of objects. For example, primitive concepts are concepts that are not defined in terms of other concepts in a given mind. Fodor often calls these ``elementary'' concepts. Similarly, primitive computations are computations which are not defined in terms of other computations in a given mind. A related sense in which ``primitive'' can be used is to pick out the basic parts from which a theory is built \parencite{samuels2002nativism,samuels2004innateness}. Here, these would be concepts which may perhaps be analyzable in terms of some other discipline (e.g.\ neuroscience, molecular biology), but which cannot be analyzed in psychological terms because they themselves are the terms in which psychology is expressed. By contrast, work on library learning often uses ``primitive'' to refer to any named term in a language \parencite{ellis2021dreamcoder,bowers2023top}. These might be complex compositions acquired through learning, but so long as that composition can be referred to by name, that name can be used in place of a composition, and thus treated as primitive.} concepts (e.g.\ \textsc{small}, \textsc{red}, \textsc{circle}, \textsc{blue}, etc.) and compositional operators (e.g.\ \textsc{and}, \textsc{or}, \textsc{not}) which can combine to form the target concept. Learning in this case is forming and testing hypotheses to see which meet some objective criterion, perhaps explaining the data well while being short in description length \parencite{feldman2000minimization,goodman2008rational,piantadosi2016logical}.

This approach to learning applies far beyond the Boolean reasoning Fodor considers. It is, for example, part of the early word learning model of \textcite{siskind1996computational}. In Siskind's model, a learner is assumed to have (innate) primitives like \textsc{cause}, \textsc{up}, \textsc{down}, \textsc{go}, and \textsc{on}. These primitives can be combined together through composition, for example building a possible hypothesis like \textsc{go($x$, up)} or \textsc{on(table)}. Compositions can themselves be further composed to create an even larger set of hypotheses, including things like \textsc{go($x$, on(table))}. As these expressions grow more complex, they begin to express familiar concepts such as \textsc{lift($x$, $y$) $=$ cause($x$, go($y$, up))}, where $x$ lifts $y$ when $x$ causes $y$ to go up. In general, such learning theories consider hypotheses coming from the set of all possible compositions of the primitives, known as the \emph{compositional closure} of the primitive operations. 
This kind of learning setup is powerful because compositional closure often defines a rich hypothesis space---perhaps capturing some corner of the breadth of human thinking---using only a small collection of operations.

The trouble, as Fodor sees it, is that this sort of learning is not actually concept learning. That is, he argues that whatever might be learned here, it is not the content of a concept which the learner previously did not possess. Instead, learners in these models use data to update \emph{beliefs} about which concepts (i.e.\ compositions) best explain the data. For example, Bayesian versions compute beliefs by combining a prior (usually favoring simple compositions) with a likelihood that quantifies how well a given hypothesis explains the observed data. As data accumulates, learners shift their distribution of belief about which composition is the best hypothesis. With enough data, these beliefs ideally converge. Critically, however, this process cannot change the support of the distribution (i.e.\ the hypotheses which are assigned probabilities), as this is, by definition, not part of statistical inference. Only the numerical scores can change, and these are assigned to predefined hypotheses.\footnote{Note, however, that many implementations use sampling techniques, which do not require one to represent the entire space at one time \parencite[e.g.][]{goodman2008rational,yang2022one}.} Fodor therefore concludes that forming and testing hypotheses simply cannot be the way in which the concept itself is learned. It can instead only be a way of updating beliefs about preexisting concepts.

The situation is similar for models whose hypotheses are neither probabilistic nor compositional in the same sense as Siskind's word learning model. Even in statistical models in which parameters are fit to data, the space of options must be ``innately'' specified beforehand by the modeler. Inference merely redistributes belief or weight among these prespecified options. For example, the structure of deep neural networks defines a parameterized space of theories (i.e.\ the possible assignments of weights in the network), each of which expresses a different hypothesis about how to produce judgments given data. Learning uses data to update weights, effectively encoding which hypotheses are good at predicting outcomes. Critically, the options for what values the weights could take on has to be there from the start. Changing the weights does not fundamentally change the space of things which can be learned by setting weights.

Fodor therefore concludes that what is traditionally called concept learning is not in fact concept learning at all. It is something else entirely, namely the process by which we establish our beliefs about preexisting concepts. Thus, what most of cognitive psychology refers to as \emph{concept learning}, Fodor refers to as \emph{belief fixation}. In his more ironic moments, he also refers to it as \emph{`concept learning'}, i.e.\ adding single quotes to show that while others might consider it true concept learning, he does not. While acknowledging the importance of belief fixation, Fodor denies that true concept learning occurs unless a mind comes to possess concepts which it previously did not possess.

In this version of the argument, Fodor identifies two cases he would count as true concept learning. The first and clearest case of concept learning is to learn a new primitive, i.e.\ a concept whose meaning cannot be expressed in terms of any composition of previously known concepts. This case is of particular interest because if primitives can be learned, they have the potential to change the total expressive power of the overall cognitive system. The second case is to learn a new composition of primitives. This case is murkier, primarily because of this statement from Fodor, ``If, in short, there are [primitive] concepts in terms of which all the others can be specified, then only the former need to be assumed to be unlearned. `Concept learning' can, to this extent be reconstructed as a process in which novel complex concepts are composed out of their previously given elements.'' What is difficult here is that Fodor seems to suggest with the first sentence that composing complex concepts is a form of genuine concept learning, while with the second sentence he identifies it with `concept learning' (note the use of ironic quotes), one of his notations for belief fixation, and therefore not genuine concept learning. This may seem like a small point, but we stress it because this ambiguity about learning over structured concepts is a point to which Fodor returns in both of the arguments discussed below.

The second major contribution \textcite{fodor1975language} makes to the discussion of radical concept nativism is the argument that it is impossible to ``learn a language whose expressive power is greater than that of a language that one already knows,'' which proceeds roughly as follows. Assuming that learning is a form of hypothesis testing, one possibility is that the target concept is a composition of simpler concepts. This case, however, patently cannot increase expressive power, because it means that the concept is expressible in terms of concepts the learner already knows. The kind of learning that can increase expressive power involves learning a new primitive concept. The problem is that if learning is a matter of forming and testing hypotheses, then learning a new primitive concept must necessarily involve proposing a hypothesis whose content expresses the content of this new primitive concept. No such hypothesis should be available, though, because if it were available, then the concept would not be primitive. It would be instead be expressed in terms of other known concepts. Fodor therefore concludes that it is impossible to learn concepts which increase the theoretical expressive power of the overall system. In analogy to a programming language like Python, no code you write in Python will allow it to express something Python \emph{could not already} express. Whatever expressive power the language of thought can eventually possess, it must possess innately.

\textcite{fodor1975language} does not strongly argue for radical concept nativism specifically. He instead lays ground work further developed in later work.
While the arguments favor a form of concept nativism, what is needed to make it radical is a clearer argument either that composing concepts is not learning or that nearly all concepts are in fact primitive.

\subsection{Version 2: Moving Toward Radical Concept Nativism}

\textcite{fodor1981present} takes a significant step toward radical concept nativism. His central concern is precisely the question of how much conceptual content is assumed to be innate in current theories of how people come to possess their concepts. In particular, Fodor articulates the positions of empiricism and nativism, broadly construed, and argues that the role for concept learning in each is severely limited, if it exists at all. For our purposes, the discussion makes three key contributions to the argument for radical concept nativism.

The first contribution Fodor makes is to describe one way that he thinks it might be possible to rescue classic theories of concept learning. Recall that in \textcite{fodor1975language}, he argued that such theories really boil down to belief fixation and so are not theories of how we come to acquire concepts at all. At the same time, he also suggested that composing complex concepts from primitives (which is what arguably happens in most concept learning experiments) \emph{might be} a form of genuine concept learning. \textcite{fodor1981present} resolves the ambiguity by proposing that complex concepts might initially be available as hypotheses only. They are explicitly not available to other parts of cognition (e.g.\ reasoning, memory, perception) and only become available after a period of belief fixation. They are thus learned because belief fixation is a rational process, and it counts as genuine concept acquisition because the concept would not itself be available to general cognition until after learning. One implication of this argument is that if the composition had been previously available to general cognition, it could not be learned, precisely because it was already available to general cognition. \textcite{fodor2008lot} later relies explicitly on this point to argue that all concept learning is impossible.

The second contribution Fodor makes is to provide an argument specifically in favor of \emph{radical} concept nativism. He reiterates that both nativist and empiricist theories of concepts require primitives and hold that primitives are available at birth or triggered by experience rather than learned. The idea of triggering is meant to imply a sub-mental ``brute-causal'' process that simply makes concepts available based on relevant experience. Such concepts are considered possessed even as they are somehow ``latent.'' The right sort of experience is required to ``trigger,'' ``activate,'' or ``manifest'' them.\footnote{While triggering is a common feature of discussions about concept nativism, there is some debate about the coherence of the idea\@. \textcite{cowie1999what}, for example, argues that triggering cannot rely on the activation of preexisting ``protoconcepts\@.'' She further argues that whether concepts are acquired through direct experience or by relying on the evidence of someone else's experience, the acquisition requires more than mere experience; it requires some sort of mediating psychological processing, often learning.} Whether available at birth or triggered, the key point is that both nativism and empiricism require sort set of innate primitives. Moreover, the space of possible concepts is completely determined by that innate endowment. As discussed above, it is not possible for learning to increase the overall theoretical expressive power of the mind by acquiring new primitives. The key question then becomes the empirical one of determining which concepts are actually primitive, therefore unlearned, and therefore innate. Fodor considers a variety of evidence both here and elsewhere \parencite{fodor1980against} and concludes that every effort to reduce lexical concepts---i.e.\ those which are not obviously composed from simpler parts---to other lexical concepts has failed.\footnote{While much of Chapter 3 in \textcite{fodor1975language} can be seen as a precursor to these arguments, the focus of the argument is sharper, and the conclusions therefore stronger, in \textcite{fodor1981present}.} He cannot find lexical concepts that reduce other than those specifically designed to be reducible (e.g.\ \textsc{prime number}). So, even if Fodor allows that concept learning through compositional construction might be possible, the claim here is that it is at least uncommon and perhaps unprecedented outside a few technical disciplines. If so, then the result is a fairly radical concept nativism: nearly all our lexical concepts are primitive and so unlearned.

Fodor considers two ways of defeating this conclusion, namely by replacing definitions with prototypes or by using a classic empiricist technique known as \emph{mental chemistry} (the details are unimportant here). He finds, however, that no one can explain how protoype-based primitive concepts compose into complex concepts, and that mental chemistry is in fact just nativism in disguise. The argument in favor of radical concept nativism stands.

The third contribution Fodor makes in this work is to give additional insight into what he means by \emph{innate} and \emph{concept} when talking about radical concept nativism. He says explicitly that, ``\emph{Roughly}, a concept is that sort of mental representation which \emph{expresses a property} and \emph{is expressed by an open sentence} [i.e.\ a proposition containing one or more free variables].'' That is, DOG expresses the property of dogness and can be expressed by the open sentence ``$X$ is a dog''. As in \textcite{fodor1975language}, he does not clearly define innateness, but gives suggestive parallelisms contrasting learning and innateness. When talking about empiricism and nativism, he says that, ``Both stories assume that primitive concepts are, in a certain sense unlearned, indeed that they are, in a certain sense, innate.'' The broader context shows that the ``certain sense'' for, say, empiricism is that, ``The range of potentially available primitive-sensory concepts is thus fixed when you specify the biological endowment of the organism, as is the range of stimulations which can occasion the availability of those concepts.'' That is, these primitive concepts are innate not in that they are directly genetically encoded or available at birth; they can be ``potentially available.'' They are instead innate in that some aspect of the organism's biology specifies both that they are possible and that certain stimuli lead to their non-rational acquisition.

One last point to note is that Fodor ends the paper seeming slightly uneasy with the idea of radical concept nativism. He says, ``I don't expect you to be convinced. I am not convinced myself.'' He instead frames the argument mostly as showing that concept learning is on uncertain footing and that some strong form of nativism may be true.

\subsection{Version 3: Radical Concept Nativism and Non-Psychological Acquisition}

\textcite{fodor2008lot} updates and clarifies the arguments of \textcite{fodor1975language} while also introducing new claims. It makes two significant contributions to the discussion of radical concept nativism, presenting both Fodor's strongest claims on the topic as well as his own attempt to escape their consequences.

The first major contribution is that Fodor presents a revised version of his earlier arguments for nativism. In explaining the need for the update he says, ``I have come to agree with my critics that there \emph{is} something wrong with the argument as \emph{LOT 1} presented it; namely that its conclusion is too weak and the offending empirical assumption---that quotidian concepts are mostly primitive---is superfluous. What I should have said is that it's true and a priori that the whole notion of concept learning is per se confused. \emph{Punkt}.'' Having reviewed the central arguments from \textcite{fodor1975language} and \textcite{fodor1981present}, he claims that, ``If a creature can think of something as green or triangular, then that creature has the concept GREEN OR TRIANGULAR,'' and that this makes it impossible to learn even complex concepts.\footnote{One difficulty here is that Fodor never explains what counts as ``thinking of something;'' he notes that he uses the term synonymously with ``bringing before the mind as such,'' but he never defines that term, either.} To see why, notice that learning \textsc{green or triangular} using hypothesis testing requires a hypothesis like ``the \textsc{green or triangular} things are the green or triangular ones.'' If so, then the target concept is necessary simply to express the hypothesis. It therefore cannot be learned as a result of hypothesis testing. The same problem that makes it impossible to learn primitive concepts also makes it impossible to learn complex concepts: traditional treatments of concept learning are such ``that learning a concept involves acquiring a belief about its identity,'' a belief which necessarily uses the concept itself.

On the face of it, this argument seems by far the strongest form of radical concept nativism Fodor proposes. It argues that all possible concepts, whether primitive or complex, simply cannot be learned. Because learning is only one species of concept acquisition, however, it leaves open that it might be possible to acquire concepts in other ways. The second major contribution \textcite{fodor2008lot} makes is to leverage this distinction between concept learning and concept acquisition to sketch a way in which concepts could be acquired without being learned. Perhaps Fodor's primary motivation in seeking out such a path is to make sense of the role experience plays in shaping the concepts available to us. He is looking for some way to avoid ``conceptual preformationism,'' the view that our concepts are like our fingers, a phenotypic expression of our genetic endowment and nothing more. He is looking for a way in which, ``Maybe experience is what bridges the gap between a genetic endowment and its phenotypic expression.''

The core of the proposal is to break concept acquisition into two phases. In the first phase, people use inductive inference to form a sort of stereotype or prototype for the concept. Fodor allows that this phase is a form of learning, but as discussed above, he dismisses prototypes as a candidate for conceptual content. The upshot is that what is learned during this phase is not itself a concept. Concepts are instead acquired during the second phase, which takes learned stereotypes and produces concepts that ``lock'' onto some property evidenced by the stereotype. The key feature of this process for Fodor is that it is not a rational or psychological process. It is instead a brute-neurological process, ``a kind of thing that our sort of brain tissue just does.'' Fodor gives very little detail as to how such a process might work but suggests that perhaps it could be thought of as relying on a space of attractors. After learning a stereotype sufficiently close to an attractor, the process would automatically spit out a new concept. As a result, no learning would be involved in actually acquiring concepts, allowing for concept acquisition without concept learning.\footnote{While Fodor argues that the initial geometry of the space of attractors must be given innately, he leaves open the possibility that this geometry might be modified over time by maturation, learning, and any number of other variables. This provides another way that learning and experience could influence concept acquisition without being involved in the acquisition itself.}

\section{The Problem to Solve}

Cognitive scientists have responded strongly to the arguments for radical concept nativism. Many believe their conclusions are ``ridiculous'' \parencite{carey2009origins} and view it as reasonable to reject arguments that
\begin{enumerate*}[(i)]
  \item affirm the innateness of all lexical concepts in use around the world;
  \item run counter to our inner experience of learning; and
  \item deny that we ever acquire or invent new concepts.
\end{enumerate*}
On the other hand, many also believe that these arguments have ``never been convincingly refuted'' \parencite{reiss2022conquer}.

One challenge in addressing arguments for radical concept nativism is identifying exactly how to go about refuting them. For example, it is unclear exactly what is at stake in the the conclusion that most lexical concepts are unlearned. The writing sometimes seems to equate ``unlearned'' with ``innate'', while at other times it carefully distinguishes learning from other forms of acquisition so that concepts can be acquired and thus not be innate. At still others, it seems to suggest that the same concept can be both acquired and innate. Beyond the question of innateness, the arguments vary in several other significant ways. They are not, for example, particularly consistent on the status of complex concepts. \textcite{fodor1975language} leaves some doubt as to whether they represent a case of belief fixation or genuine concept learning, \textcite{fodor1981present} allows that they could be learned if the mind happens to organized so as to separate candidate concepts from those used in productive thought, while \textcite{fodor2008lot} clearly argues that they cannot be learned even in principle. Nor are the arguments particularly consistent about whether unlearned concepts are acquired some other way; one says little about it \parencite{fodor1975language}, another emphasizes the way concepts could be triggered by arbitrary stimuli \parencite{fodor1981present}, and another emphasizes the need for an account that links experience and concept acquisition without resorting to learning \parencite{fodor2008lot}. They are also never all that explicit about what it means to possess a concept (though Fodor discusses it at length elsewhere), perhaps because exactly what it means to possess a concept is irrelevant in an argument which shows learners using the very concepts they are supposed to be learning. Whatever it means to possess a concept, people better possess the concepts they are actively using.

In short, the argument against learning primitive concepts is one of the few constants in arguments for radical concept nativism. Indeed, the differences in the treatment of other aspects of the problem highlight its centrality. The crux of the argument is that concept learning must evaluate hypotheses, i.e.\ beliefs about the concept's content. To do so, however, the hypothesis itself must contain some object intensionally equivalent to the concept itself. Because the arguments assume that concepts either lack internal structure or are built from other concepts, this means that the target concept must be used to express the hypothesis about what the target concept itself means. It therefore cannot be learned through hypothesis testing.

Because this argument features so prominently in the defense of radical concept nativism, refuting it remains the most straightforward path to refuting radical concept nativism as a whole. Previous counterarguments to radical concept nativism have thus focused almost exclusively on finding a way to show that primitive concepts can be learned.

\section{Previous Responses}

As just described, the arguments for radical concept nativism require a specific kind of response, one which describes both what it means philosophically to acquire a concept and how it might be psychologically possible for those conditions to obtain in human minds \parencite{laurence2002radical}. In this section, we review three types of responses. First, we look at responses which fail to meet this standard. They refute---or just deny---radical concept nativism in a way that sheds no light on how concept learning might actually work. These responses either endorse the arguments' conclusions or deny the conclusions out of hand. Second, we look at responses which attempt to meet this standard in a limited way by seeking to establish concept learning through detailed case studies. These responses often rely critically on disputed interpretations of empirical phenomena. Even granting the proposed interpretations, it is rarely clear how these arguments should generalize beyond the case study domain. Third, we look at a variety of other responses which grapple with the key issues in the arguments. Where successful, these responses indicate that concept learning is not only possible but common.

\subsection{False Starts}

Many responses to the arguments for radical concept nativism fail to refute them in a way that helps us better understand where they went wrong or how concept learning might actually occur. These responses divide into two broad categories. The first is acceptance. Some endorse the arguments' logic and maintain with \citeauthor{reiss2022conquer} that, ``There are no serious theories of learning from scratch. Learning is always a form of hypothesis confirmation that depends on some pre-existing, built-in primes, which cannot themselves be learned.'' For example, \textcite{jackendoff1992languages} says that, ``\,`learning' can consist only of creating novel combinations of primitives already innately available,'' and that, ``If I were to claim that some conceptual unit is a primitive, and it turned out to be learned, I would have no choice but to change my claim and to look for a deeper or more primitive set of basic units from which my unit could be constructed.'' Similarly, \textcite{levin1991introduction} cite Fodor as demonstrating just how much hangs on conceptual structure. On this view, whatever process we might identify as learning, it cannot create genuinely new concepts. At best, it is a search through the space of compositions. Notably, few authors embrace Fodor's (\citeyear{fodor2008lot}) final conclusion that all concepts (even the compositional ones) are innate, though \textcite{piattelli-palmarini2017fodor} is ``impressed by the cogency of [Fodor's] argument'' and goes so far as to ``plead for acceptance.\@'' He also cites the speed with which children acquire their lexicon as evidence that the concepts are waiting to be triggered or activated rather than learned. Far from refuting radical concept nativism, these responses ultimately embrace some form of it.

A second response is to flatly reject the conclusion by taking the idea that people acquire new concepts as an established empirical \emph{fact}. From this point of view, the arguments are simply ``too bizarre to take seriously'' \parencite{lakoff1987women}\@. \textcite{churchland1986neurophilosophy} writes that the idea that ``there is no such thing as real concept learning{\ldots}is sufficiently unacceptable to be a reductio of the entire mentalese hypothesis,'' an argument later seconded by \textcite{clark1994language}. To them, if something like a language of thought leads to such absurd conclusions, then ``something entirely different is going on.\@'' Our own view is that there are strong independent reasons to believe in a language of thought \parencite{fodor1975language,quilty2022best,baum2004what,kazanina2023neural,piantadosi2016four,rule2020child} and, moreover, that the arguments are not closely tied to the language of thought since they apply to many learning systems\@.

\textcite{putnam1988representation} argues that evolution could not possibly have prepared us for all possible sociocultural contexts (e.g.\ those where \textsc{perceptron} and \textsc{hygge} are relevant) and so did not do so. He writes that in order for the arguments to hold, ``Evolution would have had to be able to anticipate all the contingencies of future physical and cultural environments. Obviously it didn't and couldn't do this.\@'' Such approaches do less to solve the puzzle than to point out that the conclusion is obviously unworkable. They fail to pinpoint what is wrong with the arguments' reasoning, and thus to help explain real concept learning.

\subsection{Case Studies}

Several more detailed arguments as to how concept learning could be possible focus on case studies of particular kinds of concepts. Susan Carey (\citeyear{carey2009origins,carey2015theories}), for example, considers the case of domain-specific learning mechanisms an ``easy route to new representational primitives.\@'' In particular, she considers how the migratory indigo bunting develops its highly attuned navigational abilities. It relies on observations of the night sky; raising a bird inside an artificial planetarium with an incorrect star map can lead to permanent disorientation \parencite{emlen1967migratory}. Furthermore, without access to these observations---e.g.\ if the bird is reared without access to a star map---navigational abilities never develop. Carey highlights that indigo bunting navigation thus provides a case, just one of many available in the animal literature, where an organism acquires a genuinely novel mental representation as the result of learning. However, this case is somewhat ambiguous because indigo buntings may not learn a \emph{concept} as such---the representation is domain-specific and is not clearly part of a larger system of productive thought. Even if indigo buntings have a \textsc{north star} concept, it may be that they already have this concept prior to observing the night sky and are instead simply learning information \emph{about} the concept (i.e.\ where the star is located) which finally makes it useful in everyday inferences. Indeed, the birds may not be learning anything genuinely new and may instead just be \emph{tuning} an innate representation rather than acquiring a totally novel one.

\citeauthor{laurence2002radical} (\citeyear{laurence2002radical}; see also~\cite{margolis2011learning}) propose another possible solution based on natural kinds. They start with Fodor's theory of conceptual content: in place of definitions or prototypes, Fodor suggests that a concept represents a property based on the causal connection between that property and the presence of the concept in the mind. That is, \textsc{dog} represents dogness because when there is dogness in the world (e.g.\ an actual dog, a picture of a dog, a dog bark), the mind brings forth \textsc{dog}.\footnote{This is an oversimplified cartoon, but the nuances of the theory are unimportant for the discussion here. Two details, however, might be useful. First, the theory explicitly allows for error. While dogness should reliably token \textsc{dog}, other things might occasionally or mistakenly token \textsc{dog}, too. Second, no single piece of information about dogs is necessary to have the concept \textsc{dog}. What is necessary is the right causal relation between dogness and \textsc{dog}.} Acquiring \textsc{dog} is acquiring this causal connection. Margolis and Laurence argue that biases toward shape-based categorization \parencite{landau1988importance} and psychological essentialism \parencite{gelman2004psychological} lead a child to token a new concept when encountering a novel natural kind. That is, when children first encounter a dog, they perhaps recognize its overall shape as novel compared to other creatures. Given the bias to categorize by shape, they perhaps token a new concept for dogs, i.e.\ \textsc{dog}. This new shape-driven concept is strengthened by a bias to see members of a natural kind as sharing some deeply fundamental and \emph{essential} trait that separates them from other natural kinds. Over time, these biases guide learners toward the appropriate causal connection between dogness and \textsc{dog} and thus toward acquiring the concept itself\@. \textcite{rupert2001coining} tells a schematically similar story emphasizing philosophical issues rather than the psychological ones specific to natural kinds. While this overall approach seems promising, Margolis and Laurence consider the problem open and stress that their sketch is just an example of the broader strategy they deem necessary. They emphasize that an actual solution to the puzzle will require a much stronger defense of the theories of content and acquisition than they actually make.

Susan Carey (\citeyear{carey2009origins,carey2014learning,carey2015theories}) proposes another solution focused on natural number concepts. She argues that children learn number using a process she calls ``Quinian bootstrapping'' to form an analogy between sets and the verbal list of counting words. The bootstrapping process maps empty placeholder symbols (i.e.\ count words) to schematic representations of small sets of objects. Generalizing this mapping then leads to number concepts that exceed any innate representation and so qualify as new primitives and thus genuinely new concepts. Whether this account leads to genuine concept leraning---regardless of the empirical details---depends on how one might specify the space of possible analogies. Carey is not specific here \parencite{beck2017can}, but related proposals \parencite{buijsman2021acquiring,piantadosi2012bootstrapping,piantadosi2023algorithmic} posit that such learning (including something like the analogy) happens in a formal language whose primitives define a large space of possible outcomes. Indeed, \textcite{rey2014innate} responds that children already have the raw ability to express natural number concepts and, following \textcite{fodor1981present}, need only to trigger or manifest them. He concludes that while Carey's account shows how a conceptual system might increase its ``functioning psychological expressive power,'' arguments about the inability to increase ``semantic expressive power'' remains untouched. Rey responds similarly to the case of natural kinds. Rey's arguments, however, rely on an interpretation of innateness that ultimately appears vacuous. We discuss this point in more detail below.

\subsection{Other Approaches}

We close with a discussion of several other families of responses that are less focused on particular case studies. Many responses focus on denying individual premises, particularly that learning requires hypothesis testing. Samet and colleagues \parencite{samet1986troubles,samet1989innate} observe that many kinds of acquisition (e.g.\ catching the flu, taking a photograph) and even many kinds of learning (e.g.\ conditioning, habituation, learning food aversions or to shoot a free throw) require no hypothesis testing. They thus suggest that perhaps concept learning also occurs without hypothesis testing but give no concrete proposal as to how this might work\@. \textcite{laurence2002radical} note that imagination plays a central role in Hume's empiricism. Hume claims that imagination can decompose complex ideas into their primitive parts and recompose those parts to form fantastic concepts like \textsc{unicorn} and \textsc{dragon}. This process occurs without hypothesis testing, but as Laurence and Margolis note, it also seems to fail in the face of the argument that empirically, most concepts are not actually structured in the way Hume supposes\@. \textcite{piattelli-palmarini2017fodor}, while generally supporting radical concept nativism, says that the principles and parameters approach to language learning is so different from the standard hypothesis testing view that it may provide an example of genuine concept learning. As we observed earlier, however, parameter setting---the core learning operation on this approach---can be viewed as a kind of hypothesis testing and so is covered by existing arguments for radical concept nativism\@. \textcite{margolis2011learning} present what strike us as some of the most successful arguments against this premise. One observes that not all learning is inductive inference justified by internally rational processes. Communication, habituation, memorization, and associative learning are other forms of learning subject to different types of justification and can serve as avenues for learning concepts without hypothesis testing. Others rely on distinguishing acquisition and learning such that one can acquire and learn concepts without circularity. For example, perhaps learning occurs by revising existing concepts in such a way that learning occurs in forming the revision itself, something which necessarily happens before the final concept is even constructed. These arguments all deserve careful consideration, though most only apply in the learning of specific kinds of concepts and lack any detailed empirical evidence.

Another group of responses rely on the idea of assembling concepts from subconceptual parts. Instead of arguing that concepts can be composed from other concepts, they suggest that perhaps concepts can be composed from nonconceptual mental representations. For example, \textcite{landy2005how} argue that ``the space of possible hypotheses can in fact be expanded'' through ``primitive construction.\@'' In particular, they propose that conceptual primitives are constructed from subconceptual ``stimulus elements'' that contain perceptual information. Because primitives are constructed from subconceptual parts, their introduction counts as genuine concept learning. They also suggest that training can lead to changes in the available stimulus elements and thus expand the capacity for novel concepts\@. \textcite{mandler2008birth} claims that children use language, perceptual information, and analogy to learn genuinely new concepts. Language cues the need for new symbols, which inherit meaning from existing superordinate concepts. This meaning is then elaborated using \emph{unconceptualized} perceptual information and analogy (e.g.\ \textsc{comprehend} may be the spatial concept \textsc{taking in} alongside perceptual data about the experience of comprehension.)\@. \textcite{barsalou2005situated,barsalou2009simulation} also argues that conceptual representations are rooted in reenacting previous perceptual, motor, and introspective experience and adapted as needed to fit an individual's present situation\@. \textcite{casasanto2015all} go even further and argue that humans do not have concepts but instead \emph{conceptualize}, dynamically constructing useful knowledge structures from past experience to match the demands of each new situation. As new experiences accrue, it becomes possible to conceptualize in new ways\@. \textcite{viger2005learning} takes a somewhat different approach, presenting an argument rooted in use theories of word meanings. He argues that formally-defined words (e.g.\ \textsc{and}, \textsc{or}, \textsc{not}, \textsc{differential equation}), unlike predicates (e.g.\ \textsc{cat}, \textsc{runs}), can be defined entirely in terms of nonconceptual rules of use. If so, then learning those rules allows a formal term to be used in language and thus expands the speaker's cognitive abilities (assuming it was previously unavailable). While we are largely sympathetic with these proposals, a radical concept nativist might argue that the nonconceptual representations must still be represented somehow, such that learners must be able to represent possible concepts prior to acquiring or using them. We show how to address this concern by incorporating a general notion of subconceptual representations alongside other strategies as part of our argument against radical concept nativism.

\textcite{sterelny1989fodor} argues that conceptual role semantics are essential to rescuing concept learning. Conceptual role semantics express conceptual content in terms of the inferential relationships in which concepts participate \parencite{block1987advertisement}. One learns about mass in physics, for example, in terms of its relationship to force, acceleration, velocity, energy, momentum and so on (i.e.\ $f = ma$, $e = mv^{2}/2$, $p = mv$). The essential claim is that conceptual role makes it possible to acquire a concept without the kinds of hypotheses---i.e.\ definitions or prototypes---Fodor argues against. One simply introduces a new mental symbol and begins to describe its relationship to existing concepts. Any sort of causal relationship between the concept and what it refers to out in the world is crucially not part of acquiring the concept itself; it is a separate process. While Sterelny's focus is not on how, psychologically, concepts might be created or related to existing concepts, accounts that mirror some of the properties can be found in \textcite{piantadosi2021computational}, where an implemented model learns conceptual roles for simple domains like counting, hierarchies, and quantification. Even so, Fodor argues at length against the suitability of conceptual role semantics and related approaches as theories of conceptual content \parencite{fodor1975language,fodor1992holism}.

\textcite{weiskopf2008origins} argues for a psychological mechanism allowing individuals to token a new symbol and thereby transform an indirect representation (e.g.\ \textsc{the thing I had for lunch}) into a direct one (e.g.\ \textsc{sandwich}). Assuming you have pizza for lunch tomorrow, this new symbol has the effect of allowing you to reason about the category of sandwiches independently of the way you first encountered them. \citeauthor{weiskopf2008origins} claims that this effect actually increases expressive power because being what you ate for lunch is different from being a sandwich, so the new symbol enables you to represent a property---being a sandwich---that you hypothetically could not previously represent. Essentially, this mechanism provides a direct handle on some aspect of the world to which you previously only had indirect, contingent, descriptive access. It reduces the information required to reason about that part of the world by giving it a durable label. This seems like a plausible approach to certain families of concepts (e.g.\ natural kinds).

Finally, \textcite{deigan2022don} observes that some learners critically rely on randomness to generate hypotheses. For example, many probabilistic models of concept learning sample hypotheses from a distribution \parencite[e.g.][]{goodman2008rational}. Because this process produces hypotheses at random, selecting a particular hypothesis now does not mean that a learner will be able to reliably select that same hypothesis again in the future. It may take a very long time to randomly revisit it. Deigan argues that such models are therefore \emph{unable} to represent a hypothesis---in the sense of being able to reliably consider it as a candidate meaning---until it is sampled and given some label by which it can later be recalled. That is, he argues that concept possession depends less on the raw \emph{possibility} of representing a concept and more on being able to do so \emph{reliably}. He thus concludes such systems can learn without circularity. This argument requires randomness to play a critical role in hypothesis selection and so would not apply in cases where learning occurs deterministically. As we shall see, however, the key insight is that selecting a single hypothesis from a large space requires significant information. This insight makes an even more general argument possible.

\section{Intermission}

So far, we have discussed three different arguments from Jerry Fodor and their contributions to the case in favor of radical concept nativism. We have also reviewed a variety of responses to those arguments and found that none are widely accepted as refuting radical concept nativism in any general sense. While there are perhaps strong arguments for certain kinds of concept learning, it is far from clear that concept learning should be seen as a common part of development. In what remains, we argue against key assumptions that the arguments for radical concept nativism make about expressive power, conceptual structure, and concept possession. These assumptions are problematic because they do not reflect the reality of human cognitive development. We address each in turn with a series of observations that serve to substantially reframe what it could possibly mean to learn a concept. We then look at how they work together to refute the arguments for radical concept nativism.

One general point that is worth addressing before continuing is that discussions of concept learning and concept nativism do not always agree about what exactly concepts are or what function they serve in the mind. For example, the proposals discussed so far, both from Fodor and from others, vary widely in the sort of content which they assign to concepts. Rather than defend a particular theory of conceptual content, our position in the remainder of this paper will instead be that the exact nature of a concept is less relevant to the question of concept learning than its status as a computational object. If we assume that the mind is itself a computation, then concepts must themselves also be computations, because they are part of the mind. That is, concepts form a subclass of the computational objects which comprise the mind. Our arguments will rely heavily on the idea that concepts are computations, but will otherwise remain agnostic as to the exact nature of conceptual content.

\section{The limits of expressive power} \label{sec:express}

Primitive concepts play a central role in existing arguments for radical concept nativism. If most lexical concepts are in fact primitive, then acquiring a new concept allows a mind to express something which it could not have expressed previously. Coming to acquire new primitives is thus important because it expands a mind's expressive power. In this section, we argue that it is impossible in principle for human minds to expand their expressive power---through learning or any other kind of acquisition---and thus that the picture of concept learning on which radical concept nativism relies does not reflect human cognition. The implication is that whatever people might be doing when they begin to make use of concepts they have not previously used, it cannot be explained in terms of the kinds of concept acquisition assumed in arguments for radical concept nativism. For people, that kind of concept acquisition is not merely impossible, it is incoherent.

Backing up slightly, one of the key arguments in Fodor's (\citeyear{fodor1981present}) defense of radical concept nativism is that nearly all concepts are primitive \parencite{fodor1980against}. The few exceptions primarily involve obviously phrasal concepts (e.g.\ \textsc{quick brown fox}) and technical concepts explicitly designed to be definable (e.g.\ \textsc{multiplication}). Because most concepts are primitive, there is no way to express them in terms of other concepts. If you want to be able to think about a fox, you must have the concept \textsc{fox}. On this view, each new primitive a mind acquires---whether through learning or otherwise---increases its expressive power. Acquiring \textsc{fox} is what makes it possible to think about foxes.

One key assumption of this approach is that the mind's expressive power \emph{can} in fact be expanded. Readers might then suspect that there are typically interesting concepts outside of every learning model's compositional closure and which are therefore not part of the hypothesis space. This is indeed exactly what you should expect if \textsc{fox} is primitive: acquiring it would increase expressive power by making new thoughts possible, namely thoughts about foxes. Moreover, some systems show this kind of behavior: if we consider a conceptual system with limited computational power (e.g.\ one equivalent to predicate logic), there are some concepts it can express and others it cannot. In these systems, it is meaningful to ask how adding a primitive (e.g.\ universal quantification) might expand the compositional closure and thereby increase expressive power.

Remarkably, not all systems can be expanded in this way. In particular, the class of \emph{Turing-universal} systems can express all possible computations. For example, there are individual Turing machines which can simulate every other Turing machine \parencite{turing1936computable}, and thus run every known computation. The key takeaway here is that Turing-universal systems can already express \emph{every possible computation}: assuming the Church-Turing thesis is correct, their expressive power simply cannot be expanded.

Importantly, many---starting with \textcite{turing1936computable}---have suggested that human minds are precisely these kinds of computational systems, capable of expressing any possible computation \parencite{baum2004what,rule2020child,hutter2004universal,chater2007ideal,piantadosi2016four}. Indeed, Turing initially developed the Turing machine as a formal model of a human ``computer'', which referred at the time to a mathematician working with pencil and paper (the source of a Turing Machine's tape). Many of his considerations in the design of the machine, such as the use of a finite number of symbols, are thus dictated by his intuitions about what a human might be capable of doing without being overwhelmed or confused. It is thus only \emph{after} he has described what he believes to be an abstract model of this human computer---frequently referred to in the text as ``he''---that Turing then suggests, ``We may now construct a machine to do the work of this computer.\@'' One reason to believe that humans are Turing-universal, then, is that Turing-universal computation was designed from the start to model human computational abilities \parencite{copeland2000narrow,copeland2013turing}.\footnote{A related argument \parencite[see, e.g.][]{baum2004what} establishes that minds are no more than Turing-universal by working backward from Turing machines to the mind. It observes that the brain can be in only a finite (but very large) number of states, such that it is theoretically possible to simulate a brain using a Turing machine. If so, then if the brain gives rise to the mind, the mind itself can certainly be simulated by a Turing machine.}

\begin{figure}[t]
  \centering
  \begin{subfigure}{0.60\textwidth}
    \adjustbox{raise=1.2em}{\includegraphics[width=\textwidth]{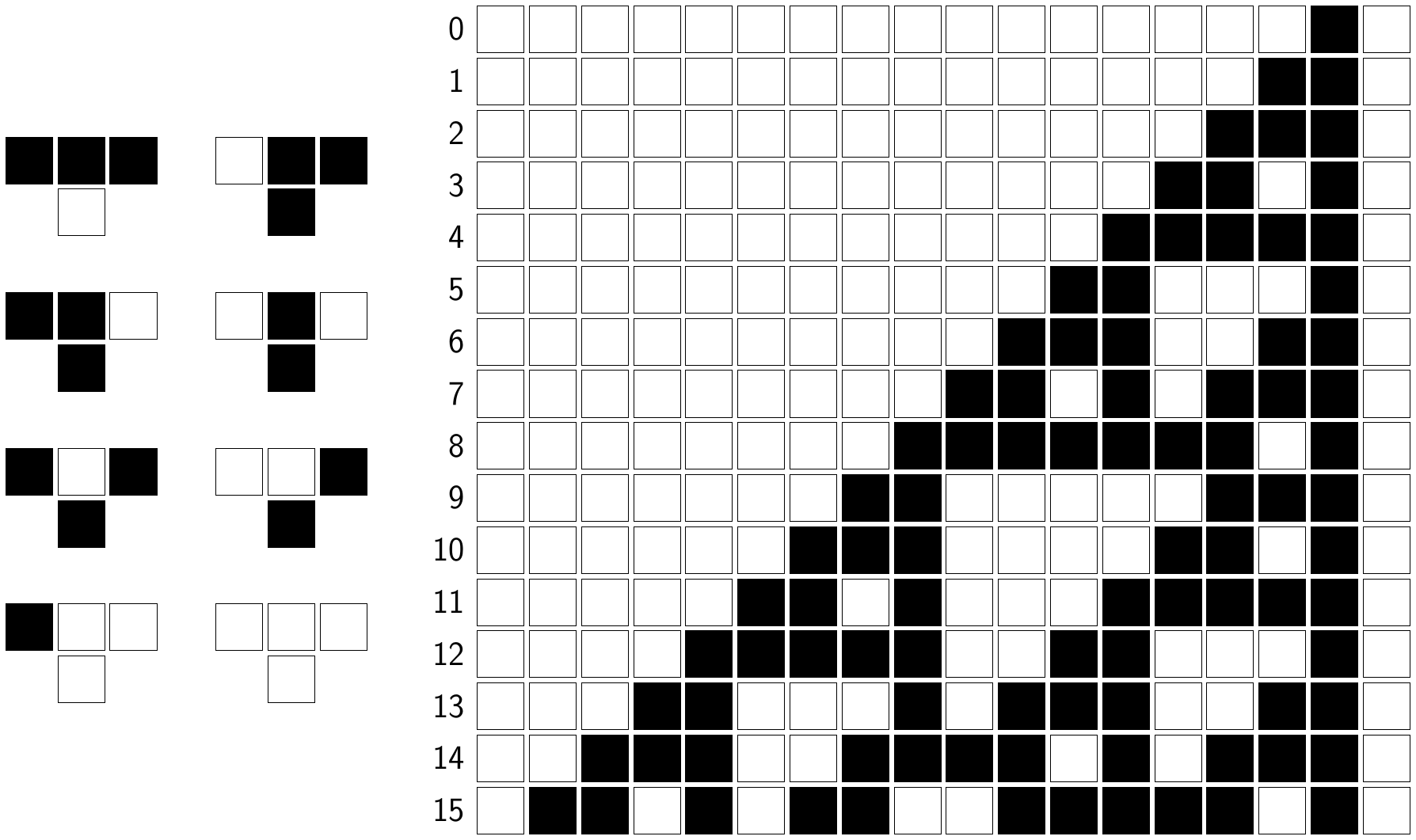}}
    \caption{}\label{fig:rule110}
  \end{subfigure}
  \hfill
  \begin{subfigure}{0.38\textwidth}
    \includegraphics[width=\textwidth]{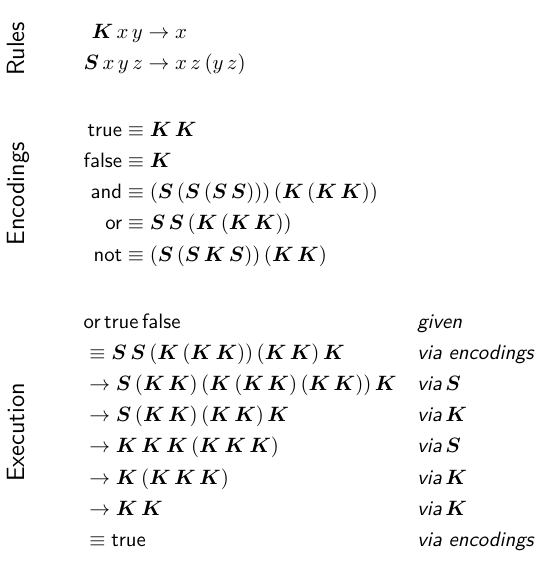}
    \caption{}\label{fig:sk}
  \end{subfigure}
    \caption{Simple Turing-universal systems whose compositional closure cannot be expanded, even in principle. (a) Rule 110 \parencite{wolfram2002new} uses two primitives, $\square$ (white) and $\blacksquare$ (black), arranged in sequence. Different sequences express different computations. Eight rules (left) describe how to advance the computation. For example, the top-leftmost rule indicates that a black with two black neighbors should become white; An example evaluation (right) demonstrates 16 execution steps (rows), progressing from top to bottom. The entire sequence updates simultaneously. (b) SK combinatory logic \parencite{schoenfinkel1924uber,curry1930grundlagen} also uses two primitives, $\bm{S}$ and $\bm{K}$. Each represents a function following a simple rule (top). More complex behaviors and data structures can be encoded using these simple functions (middle). Execution proceeds simply by rewriting terms according to the rules and any defined encodings (bottom).}\label{fig:tcSystems}
\end{figure}

Another reason to believe that minds are Turing-universal is that many people master provably Turing-universal formalisms. People learn about Turing machines (including universal Turing machines), combinatory logic, lambda calculus, Rule 110, and the many mainstream programming languages which are Turing complete. Even more strongly, they can reason about the behavior of these systems. Indeed, some of these systems are very simple. For example, some universal Turing machines use fewer than two dozen rules \parencite{rogozhin1996small}. Extremely simple systems of rules such as Rule 110 (Figure~\ref{fig:rule110}) and even carefully initialized collections of colliding Newtonian particles \parencite{fredkin1982conservative} can also express arbitrary computations, often in surprising ways \parencite{wolfram2002new}. The broader point, however, is that if humans were not themselves Turing-universal, they provably could not use such systems or reason about their behavior.

To clarify, in calling humans Turing-universal, we do not mean simply that all humans have the ability to mechanically follow the rules of a particular Turing-universal system (e.g. Rule 110), though many people demonstrably can do so. Nor are we claiming that the human conceptual system is Turing-complete and can token a concept for any computation. While this seems to be true---people can learn lambda calculus or combinatory logic and token a concept for seemingly arbitrary expressions in those systems---it is not again not the most interesting sense in which people are Turing-universal. Instead, what we mean with this claim is that there is no theoretical limit on the computations that human minds can express as subcomputations. Human minds are instead capable of synthesizing, executing, and even analyzing effectively any computation. Some may be identifiable with concepts, but many others might be better labeled, among other things, as: beliefs; learning mechanisms; motor programs; perceptual interpreters; emotions; desires; or goals. In this sense, concepts are just part of the broader cognitive landscape.

If humans are Turing-universal in this sense, then inductive learning's impact on expressive power loses significance for theories of human cognition. At least for people, the question of increasing expressive power becomes moot because there is no expressive power left to gain. This is one of Turing's most powerful ideas---indeed it may be the most important idea in the history of computation. Whatever it might mean to acquire a concept, it cannot mean coming to express a concept which could not previously be expressed. There are no such concepts (that can be realized by a computing system).

At this point, one might object that Turing's arguments about universal computation apply to general-recursive functions over numbers. Even if humans are computationally complete in this sense, it is unclear what relevance it has for psychological questions such as whether concept learning is possible. However, if we assume that concepts are part of the computation that is the mind, then they must also be computations. In that case, every concept can be identified with some general-recursive function. That is part of what it means to treat the mind computationally. As a result, being Turing-universal means that a system can express all possible general-recursive functions and thus all possible concepts. This argument does not depend on any particular theory of conceptual content but instead on the more basic hypothesis that the mind itself is a computation.

One might also object that people are bounded in their cognition and so cannot actually be Turing-universal. The idea here is that learners with Turing-universal primitives but limited resources (e.g.\ finite memory and time) become incapable of expressing some computations \parencite[the hierarchy theorems;][]{sipser2012introduction}. Thus, even if we assume that human minds use Turing-universal primitives and so have asymptotically maximal expressive power, changes in cognitive limits could lead to changes in practical expressive power for actual humans\@. \textcite{fodor1975language} accepts that increasing these limits may allow minds to express new concepts but sees such changes as unlikely to be caused by learning. Moreover, he argues that such practical constraints are theoretically uninteresting and explicitly frames his argument around theoretical minds with unlimited resources.

Combining unlimited computational resources and Turing-universal primitives, however, is enough to refute the arguments for radical concept nativism. A mind with Turing-universal primitives and unlimited resources can express every possible computation, and therefore every possible concept. It is incoherent to suggest that it might be possible, even hypothetically, to acquire concepts that could increase such a mind's expressive power. This observation does not refute any specific premise in the arguments for radical concept nativism. It instead argues against the understanding of concept acquisition which those arguments presuppose, i.e.\ that people regularly extend their theoretical expressive power by acquiring new concepts. Given unlimited computational resources and Turing-complete primitives, concept acquisition simply cannot work this way; some other explanation is needed.

It is, however, a mistake to assume unlimited computational resources if the goal is to understand whether actual humans learn concepts. Actual humans operate under stringent computational limits \parencite{van2008tractable,van2019cognition,lieder2020resource}. Empirically, these limits differ dramatically across both species and development \parencite{cantlon2024information}, and so are likely to play a key role in explaining actual concept acquisition. These limits also make genuine concept learning possible through chunking \parencite{miller1956magical,gobet2001chunking}. For example, if a memory-bounded mind can compose no more than five symbols at once, it cannot naively express computations requiring six symbols. They are theoretically inexpressible given the resource bounds. The hypothesis space is limited to compositions of up to five primitives. Imagine, however, that this mind has a chunking mechanism. In particular, imagine that it forms two three-primitive chunks and tokens new symbols to name them. If this chunking operation is a discrete event with a definite beginning and end, then the hypothesis space available before chunking contains fewer hypotheses than the space available after chunking. To see why, consider that the mind can now compose not just the initial primitives but also the symbols referring to the two named chunks. That means that the mind could compose these two new symbols to express a concept that would naively require six primitives. Before the chunking operation, this concept was inexpressible and so outside the hypothesis space; it only became expressible through the creation and naming of the chunks. Chunking can thus increase the theoretical expressive power of a resource-bound mind. Recent computational models show how Bayesian inductive inference can use this sort of chunking to increase expressive power over time for resource-bound agents, even when using Turing-universal primitives \parencite{ellis2021dreamcoder,bowers2023top}.

Concept learning through chunking thus provides another way to meaningfully refute radical concept nativism. It shows how actual resource-bound minds could use rational means to come to possess concepts they did not previously possess and could not even express. By learning a chain of chunks that build upon one another, a mind can come to represent a complex concept which cannot be represented in terms of the original primitives given practical resource limitations\@. \textcite{fodor1975language} grants essentially this same point when talking about the benefits of learning a natural language. He ultimately concludes in favor of radical concept nativism, however, by considering situations in which people do not actually find themselves, i.e.\ with unlimited computational resources and primitives of limited expressive power. Instead, human minds operate with limited computational resources but maximally expressive primitives. It thus seems likely that in practice, people can learn concepts which they could not express without chunking and so increase their practical expressive power in a manner that amounts to real concept learning.

The general point here is then that the presence of Turing-universal primitives provides a path to understanding one problem with arguments for radical concept nativism. Combining universal primitives and unlimited computational resources shows that the vision of concept acquisition undergirding these arguments cannot apply to humans. Combining universal primitives with limited computational resources enables chunking to straightforwardly increase expressive power.

Finally, it is worth noting that \textcite{turing1950computing} actually considered and dismissed a version of the argument that fixed expressive power implies radical forms of nativism. Writing about the potential for an artificial intelligence, he says that, ``The idea of a learning machine may appear paradoxical to some readers. How can the rules of operation of the machine change? They should describe completely how the machine will react whatever its history might be, whatever changes it might undergo. The rules are thus quite time-invariant.'' While expressed differently, Turing's observation is, in a nutshell, identical to Fodor's. The theoretical expressive capacity of any cognitive system is predetermined. The architecture of such a system---whether human or machine---must describe, from its inception, how it will respond to anything it could possibly encounter. If all possible responses are given ahead of time, then it certainly seems reasonable to conclude that no real learning can take place. Indeed, this is Fodor's conclusion. Turing, however, argues that, ``The explanation of the paradox is that the rules which get changed in the learning process are of a rather less pretentious kind, claiming only an ephemeral validity.'' While some rules must indeed be fixed, others can be learned. These more ``ephemeral'' aspects of the mind are acquired in the sense of being a part of thought which, while possible, had not previously been actually in use. This sense is the one in which Turing conceived learning while originally developing the computational vision of mind.

\section{Conceptual structure as computational structure} \label{sec:structure}

In this section, we review Fodor's (\citeyear{fodor1981present}) argument that concepts generally lack internal structure, a central claim in his argument for radical concept nativism in that work. We then give a very simple computational counterargument which shows that every possible concept must be definable in a way that assigns it internal structure. In doing so, we provide another avenue to rebutting perhaps the most significant argument against concept learning. We close by considering how our approach affects the later and more radical claim that even complex concepts cannot be learned \parencite{fodor2008lot}.

We begin with the claim that concepts generally lack internal structure. The claim is that, after categorizing concepts according to whether they are primitive (i.e.\ they are \emph{not} composed of simpler parts and therefore have \emph{no} internal structure) or complex (i.e.\ they are composed of simpler parts and have internal structure), almost all lexical concepts will end up as primitives. The argument for this claim combines two premises:
\begin{enumerate*}[(i)]
  \item concepts generally cannot be defined in terms of other concepts; and
  \item other attempts at providing internal structure fail to provide the systematic compositionality required to explain thought.
\end{enumerate*}
To support premise (i), Fodor and colleagues considered projects from throughout linguistics, philosophy, and psychology all aimed at showing that at least some significant body of concepts could be defined in terms of other concepts \parencite{fodor1981present,fodor1980against}. In each case, he argued that the projects failed. While not conclusive, the cumulative failures provide strong inductive evidence against the idea that concepts are generally defined in terms of simpler concepts. To support premise (ii), Fodor again considered theories proposing to give concepts internal but non-definitional structure by way of things like mental images, prototypes, definitions-in-use, and inferential role semantics. In each case, Fodor finds difficulties, primarily that these structures fail to compose in the way concepts must to explain productive thought. For example, the prototypes of \textsc{brown} and \textsc{cow} do not compose to produce the prototype of \textsc{brown cow}. Taking the two premises together, Fodor concludes that except for a small number deliberately constructed to have definitions, concepts generally lack internal structure. This ultimately empirical premise is central to Fodor's (\citeyear{fodor1981present}) argument. It would collapse if most concepts turn out to have internal structure.

We generally agree that the empirical search for concepts which can be convincingly defined in terms of simpler concepts has had little success.\footnote{We remain agnostic as to exactly how many concepts contain simpler concepts as part of their internal structure. Nothing in our argument hinges on this still unresolved empirical question.} We also agree that just about every theory of conceptual content faces serious challenges, including the alternative sorts of internal structure he considers, although recent work on vector-based models mitigates some of these concerns \parencite{piantadosi2024concepts}.

Taken together, however, these facts do not imply that concepts have \emph{no} internal structure. Our argument does not rely on the particular details of any one theory of conceptual content. It instead begins with the observation that \textcite{fodor1975language} saw the language of thought as ``the medium for the computations underlying cognitive processes.'' He notes that, ``while I have argued for a language of thought, what I have really shown is at best that there is a language of computation.'' As such, the primitives of the language of thought should be understood to be the primitives of the computational medium underlying cognition.
The critical point gained by starting here is that there is no a priori reason to suppose that cognitive primitives are concepts, beliefs, or anything familiar to the present vocabulary of psychology. They are instead the primitives of whatever computational system is actually used to express a mind.

For example, the primitives of modern digital computers are bits and the set of operators which the processor uses to manipulate them. These are far removed from the objects of typical concern such as operating systems, applications, and files. Indeed, while files and applications are primitive in some respects (e.g.\ a file is not typically composed of simpler files), they are not at all primitive in other respects. For instance, a text file containing just the string ``Hello, World!'' is actually a complex data structure containing various metadata and a list of locations where the bitstring representing the content of the file is actually stored on disk. The point is that, even if the phenomena of interest deal with files and applications, operations on bits are the ultimate primitives necessary to describe the set of things which a modern digital computer might do. While we can describe some phenomena in high-level terms (e.g.\ composing a poem can be broken down into the operations of opening a file, drafting the poem, and saving the file), other questions depend on these much lower-level operators. Similarly, while we can characterize some cognitive phenomena at the level of beliefs and concepts, others may require a different approach. In particular, questions about which computations the mind can represent or execute may depend not on convenient high-level structures but instead on the simpler structures from which they are composed, or some intermediate level of description. As a result, concepts, beliefs, and other parts of the mind may turn out to be expressed as complex compositions of much simpler computations.

To show that concepts and other mental objects can in fact be expressed as compositions of simpler parts, we turn to computer science. Some Turing-universal systems---including Rule 110 but also lambda calculus \parencite{church1932set} and combinatory logic \parencite{schoenfinkel1924uber,curry1930grundlagen}---are radically compositional, meaning that they specify a set of primitives and express \emph{all} computations as compositions of those primitives. For example, consider the black and white squares of Rule 110 (Figure~\ref{fig:rule110}). No computation can be performed with a tape containing a single square. At least 3 squares are required before any of the system's rules apply, and the exact rule which applies depends on how those three squares are arranged. Rule 110 is radically compositional. Moreover, its primitives are not arithmetic expressions or Boolean formulae. Those black and white squares exist at a level below arithmetic and logic. They can be composed, however, into patterns that express all possible arithmetic expressions and Boolean formulae, as well as all other possible computations. This holds for all radically compositional Turing-universal systems---we give a small example for combinatory logic in Figure~\ref{fig:sk}. Such systems are widely used and form the basis for many modern programming languages \parencite{cardone2006history,hudak2007history,abelson1996structure}. Moreover, there happen to be methods for constructing such systems which show that there are unboundedly many distinct ways of setting them up, each with a single primitive \parencite{goldberg2004construction}. As a result, every possible computational object can be described in a way that gives it internal structure. If it happens to be primitive in one system, it will turn out to be compositional in another. Similar arguments also apply to certain classes of deep neural networks which can represent any possible computation using the internal compositional structure of vectors and matrices.

Because they express all possible computations compositionally---including all possible mental representations---radically compositional systems also serve as a counterargument to the claim that most lexical concepts lack internal structure \parencite[see][for a counterargument from linguistics]{levinson2003language}. They show that even if concepts do not decompose into simpler \emph{concepts}, they must decompose into simpler \emph{computations}. For example, an ASCII character is not composed of simpler characters. It is composed of a string of bits. The representation can be decomposed into simpler parts which are not themselves the same kind of thing as the object they compose. The same is true for concepts. If \textsc{dachshund} is really a computational object, then some pattern of black and white squares can be used to express it in Rule 110. Along those same lines, it may be helpful to observe that concepts are not neural primitives. Whatever computation implements \textsc{dachshund} is expressed in the brain in terms of simpler computational units involving neural firing or perhaps, even more granularly, the biochemistry of individual neurons. Whether we use Rule 110 or biological neurons, individual concepts must be describable in ways that assign them internal computational structure.

Another way to see that concepts must have some internal structure is to note that without it, there is no way to tell them apart. Consider that physical atoms are not themselves atomic in the sense of being indivisible. They can be broken into protons and neutrons, with different atoms containing different numbers of these subatomic particles. It is precisely these differences in internal structure that make hydrogen, oxygen, and carbon different from one another. In the same way, even if most lexical concepts are atomic in the sense that they cannot be defined in terms of simpler concepts, they must still have enough internal structure for the mind to individuate one from another.

The point here is just that in order to tell two objects apart, they must be different in some way. For example, say that the goal is to distinguish \textsc{dog} from \textsc{cat}. They could be distinguished based on one of two kinds of differences. First, the differences between \textsc{dog} and \textsc{cat} might be internal to the concepts themselves. If so, then the objects have some sort of internal structure that distinguishes one from the other. Second, differences between concepts might be external to the concepts themselves. That is, \textsc{dog} and \textsc{cat} might themselves be identical, and it is only some metadata associated with them (e.g.\ a label, or the memory location where they are stored) which differentiates one from the other. In this case, however, \textsc{dog} and \textsc{cat} are not themselves actually distinct. They are identical and thus are not actually distinct representations of distinct concepts. It is only when considered as part of a larger structure (e.g.\ in conjunction with their label or memory location) that they can come to represent distinct computations in the broader cognitive system. However, these larger structures, by definition, have complex internal structure. That is, two identical atoms cannot be distinguished and so cannot represent different computations, concepts or otherwise. Some sort of internal structure is necessary.

To be clear, our claim is not that concepts can be defined in terms of other concepts, nor that they gain structure from things like mental images or prototypes. It is instead that, because they are computations, they must have computational structure which can be described in terms of other (simpler) computations. We make no strong claims about exactly what these simpler computations are. Our argument does not require such a claim.\footnote{Our own view is that at least some of these simpler computations are likely to be subconceptual, sitting below concepts much in the same way that the $S$ and $K$ of combinatory logic or the black and white squares of Rule 110 sit below just about every useful computation one might want to implement. The argument, however, does not hinge on this being the case for all concepts.} It is instead rooted in the fact that they can be described by radically compositional and computationally universal systems.

Thus, we think that the proposals we discussed earlier which seek to rescue learning via the composition of subconceptual parts were definitely on the right track. The main change we are proposing here is that, in light of humans' incredible computational sophistication, we can posit very simple yet general purpose computational primitives at work in the mind. The exact nature of these primitives remains an important empirical question, but the question of whether subconceptual parts can be used to express concepts (and any other cognitive structure) is already definitively answerable in a way that does not depend on empirical details.

One final point that is worth discussing here is how concepts compose under this view. The compositional nature of concepts is central to discussions about radical concept nativism. In particular, the fact that many learnable representations (e.g.\ prototypes, mental imagery) fail to compose in the ways we expect concepts to compose is one reason to exclude them as possible substrates for concepts. Doing so further strengthens the inductive evidence against concept learning. Our proposal, however, is not subject to this critique. If concepts have internal computational structure, and the mind represents concepts by representing that structure directly (i.e.\ in terms of compositions of subconceptual parts), then composition takes place naturally by applying one concept to another. Consider, for example, that mathematical functions compose by operating over the internal (set-based) structure of functions. Similarly, subroutines in a programming language compose by operating over the internal sequence of computational steps associated with each subroutine. In the same way, the proposal in this section allows for concepts to compose in the correct ways because of how the internal computational structures of the composed concepts behave when one is applied to the other. That is, it is the internal structure of \textsc{sky} acting on \textsc{blue}, and vice versa, that allows them to compose into \textsc{blue sky} and \textsc{sky blue} and which allows those complex concepts to represent distinct cognitive structures, i.e.\ distinct computations with distinct roles in the mind.

As in the previous section, the claims in this section alone are enough to refute Fodor's (\citeyear{fodor1981present}) argument. We show that concepts, because they are computations, can be described as having internal structure. Showing that concepts lack internal conceptual structure fails to take into account what the argument for the language of thought actually shows, namely that the mind relies on some collection of primitive computations. This point is profound and one we are working to bring back to the center of the conversation, because it makes it very likely that concepts do in fact have internal computational structure. If the mind can use something other than concepts as primitives, then concepts can be constructed from simpler parts, and one of the argument's premises no longer holds. The argument collapses.

One might object here that it is possible to describe concepts in terms of subconceptual parts, but that doing so has no real bearing on Fodor's concerns. Fodor's (\citeyear{fodor1981present}) argument in particular is framed as an argument against classical empiricism, the idea that concepts are learned from a small set of primitive (and generally sensorimotor) concepts. By proposing that minds build concepts not from other concepts but from subconceptual parts, we do nothing to defend this form of empiricism, i.e.\  learning by hypothesis testing over a space defined by a small set of sensorimotor primitives. Our goal in this paper, however, is not to save this version of empiricism. Fodor's (\citeyear{fodor1975language}) basic point was that we need something more than just behaviorism or neuroscience to explain psychology. We need to posit some level of computational structure, some sort of language in which cognitive computations take place. This language operates over computations, not over concepts as such. It is thus entirely possible to consider minds that learn concepts by testing hypotheses composed of subconceptual parts. Doing so is entirely relevant to the main goals of this paper, i.e.\ showing that the arguments for radical concept nativism are computationally untenable while also proposing a way in concept learning could be commonplace.

To close this section, we consider the objection that \textcite{fodor2008lot} extended his defense of nativism to preclude the possibility of learning even complex concepts. The basic idea here is that even for complex concepts, hypothesis testing requires that the content of the concept appear in the hypothesis, meaning it cannot actually be learned as a result of hypothesis testing. If so, our counterargument accomplishes nothing, because no concept of any sort can be learned in principle. We actually agree with this objection. The 2008 argument was explicitly motivated by a desire to excise the empirical assumption that most lexical concepts are primitive. We should not expect a counterargument to that assumption to defeat an argument in which it does not appear. In the \emph{Discussion}, we offer an alternative view of concept learning building on the ideas in this section and its immediate neighbors and respond more fully to Fodor's (\citeyear{fodor2008lot}) argument.

\section{A new view of concept possession} \label{sec:possess}

All three of Fodor's arguments for radical concept nativism claim that it is impossible via learning to come to possess a new primitive concept and that they must then be, in some important sense, innate. At the same time, Fodor (\citeyear{fodor1981present,fodor2008lot}) argues that (primitive) concepts are acquired without being learned. He does so in part to argue that they are, in another sense, not innate. This use of multiple senses of innateness appears throughout Fodor's writing on nativism and has led to a number of views about exactly what kind of \emph{nativism} is at stake in radical concept nativism.

\textcite{fodor1975language} seems to explicitly say that concepts are innate if they are unlearned, and many have subsequently endorsed that as Fodor's view.
Depending on the exact framing of the discussion, any concept which is acquired without learning could be considered innate. This would include concepts acquired doing things like taking Latin pills or being hit on the head. That seems undesirable, and \textcite{fodor1975language} takes great care to draw distinctions between innateness and concept acquisition precisely to preclude such examples. So, another option is to suggest that what Fodor means by calling a concept innate is that it is not acquired through experience, and \textcite{fodor2008lot} says that ``an innate concept is one the acquisition of which is independent of experience.\@'' \textcite{fodor1981present}, however, suggests that concepts are innate even if they are only potentially available and certain experiences are required to trigger their non-rational acquisition. Indeed, much of Fodor's writing about nativism is an attempt to explain how experience shapes the conceptual repertoire. Statements like these could lead to the view that what Fodor means by ``innate concepts'' are those which come to be possessed as the result of a non-psychological process.

By contrast, \textcite{rey2014innate} appears to read Fodor as taking a much stronger view, namely that any concept which can theoretically be acquired (i.e.\ anything in the primitives' compositional closure) is innately possessed. What happens in what we have been calling concept acquisition is that these innately possessed concepts becomes ``manifest'' and as a result, available to general cognition. One difficulty Rey identifies with this view, however, is in explaining why Fodor places such emphasis on the idea that even this ``manifestation'' cannot be the result of learning. Another challenge that we see is that equating expressive power and innateness transforms the discussion of radical concept nativism, in the terminology of \parencite{chalmers2011verbal}, a purely verbal dispute. It transforms the argument about concept nativism so that it hinges critically on a disagreement about the semantics of ``innateness'' and ``possession'' and nothing else---certainly nothing empirical. We can follow Chalmer's method for detecting verbal disputes, by banning the use of these terms and seeing how the claim changes. \textcite{carey2014learning} notes that doing so makes concept possession ``unproblematic, but empty.\@'' In particular, ``It has the consequence that even extreme anti-nativists would have to agree that all concepts that are ever manifest are possessed at birth. Nobody would ever deny that an actual manifest concept had the potential to be the output of some developmental process. This is bizarre terminology, and for this reason, psychologists do not talk about ‘concept possession’ with this meaning.\@''
It thus seems unlikely that Fodor consistently equated concept possession and expressive power and unhelpful to proceed as if the debate over radical concept nativism is primarily a debate about expressive power. It is not.\footnote{While we find it unhelpful to think about concept possession in terms of expressive power, our views do share something in common with Rey's (\citeyear{rey2014innate}) ``ecumenical'' conclusions, namely that the question of expressive power is separate from the question of learning. As we shall see, learning is about selecting, from among all possible computations, those which are most useful to the current explanatory project. Picking out any particular computation requires information, and it is our argument here that the source of this information determines the degree to which it is best described as learned or previously possessed.}

The broader point is that there is no clear consensus on precisely what Fodor means by an ``innate concept.\@'' It even seems likely that he perhaps had multiple competing views of innateness in mind. This is not surprising given the relatively fraught status of innateness in the study of cognition, and throughout the sciences more broadly. The term is sometimes used to reference concepts or traits which are somehow encoded in human DNA, or are actually available for use at birth, or which tend to occur despite environmental variation, or which are the result of Darwinian adaptations, and so on \parencite{bateson2007innate}. Another part of the challenge here is that concepts are far from the only aspects of cognition which may be innate: so, too, can aspects of our cognitive architecture and developmental timeline \parencite{elman1996rethinking}. Indeed, the question of exactly what it means to be innate has been difficult to answer \parencite{gross2012innateness}, and proposed answers seem to entangle related concepts \parencite{griffiths2002what} in a way that essentially mimics folkbiology \parencite{machery2019scientists}.

The upshot is that the core defenses of radical concept nativism never unambiguously state what it means for a concept to be ``innate.\@'' They instead rely on conventional usage, which allows readers to inject their own intuitions into the argument and interpret it as they will. Because ``innate'' is such a troubled term, the result is that it is difficult, if not impossible, to determine precisely what was intended and thus precisely what the argument demonstrates.

In the rest of this section, we will not attempt to disentangle the various senses of innateness which Fodor's arguments might use. We instead suggest that, given its troubled status, it is better to reframe the discussion so that it does not rely on precisely what Fodor meant. In particular, we observe that Fodor relied on binary notions of innateness and concept possession which are insufficiently rich to explain human concept learning. We then argue that concept possession is instead best thought of as a graded value and introduce a concrete measure for it which is rooted in information theory. We show how this new view of concept possession transforms debates about nativism and empiricism into what is a more unified discussion about how minds acquire the information required to represent their concepts.

We start by observing that whatever particular sense of innateness Fodor's arguments use at a given time, they make an important implicit assumption. In particular, they assume that innateness is a binary value. A concept is either innate, or it is not. That is, the exact rule for determining whether a concept is innate might differ from use to use, but the underlying assumption is that it is at least possible to determine whether a given concept is in fact innate or not under any given rule. It is related to another unstated assumption, namely that concept possession itself is a binary value, such that at any given time, a mind either possesses a concept or it does not. These two assumptions are intimately connected in that most ways to construe innateness will eventually rely on concept possession: e.g.\ perhaps a concept is innate if you possess it at birth, or come to possess it independently of experience, or if it comes to be possessed after some non-rational trigger.

Binary notions of concept possession, however, are too impoverished to explain human concept learning. To see why, consider humans' cognitive flexibility. People can respond to seemingly infinitely many situations with a diverse repertoire of responses that draw on a vast array of concepts. The same stimulus might lead to many different thoughts, depending on one's current intentions and mental state. As such, the exact concept a mind uses at any given moment must be selected from a range of options. Selecting one option from among many requires information, namely the information required to specify that particular concept rather than any of the alternatives. Without this information, no concept can be selected; even drawing one at random requires rolling a die, flipping a coin, or some other mechanism of injecting enough information to pick one concept from among many. The alternative would be that individuals would find themselves stuck in a sort of cognitive catatonia. They would be unable to select among multiple concepts. Their conceptual life would be vacuous and completely unresponsive to the flow of information around them.

Combining this idea of cognitive flexibility with the fact that humans are already likely capable of expressing any possible computation (and therefore any possible concept) and that all these computations (and concepts) can at least theoretically be expressed in terms of simpler computational parts, the result is a new view of concept possession. Given the right information, any concept can be selected, but no individual concept can be selected without at least some information picking it out from among the sea of alternatives. That is, no concept is entirely unreachable and therefore completely lacking, and no concept is immediately available and therefore possessed absolutely. Moreover, there is no way that all of the unboundedly large set of possible concepts could be explicitly represented simultaneously.\footnote{This is another reason to believe that concepts and other mental representations are composed of simpler computational parts. There's simply no way they could all be stored explicitly and atomically in finite memory.} Instead, information is required to select which of many possible concepts should be represented. This process is an active one that requires constructing or selecting a concept by extracting information from the environment in a context-dependent way.

We thus propose that instead of a binary measure, concept possession can be seen formally as depending on the amount of information required to select a concept relative to its total information content (whatever that content might turn out to be). To motivate our view, it will help to consider a family of different word processing programs. Each will support full keyboard input---i.e.\ each supports the same primitive set of characters and a means for composing them into arbitrary strings---and so has an equivalent theoretical expressive power. However, most will \emph{also} provide additional buttons supporting other kinds of operations. The objects in these examples are strings of text. Strings are useful because they are concrete, fairly intuitive, and their information content can be estimated straightforwardly. We are not claiming that concepts are strings or that the concept of any particular string of text is identical with the string itself. The point is to build some intuitions about information with familiar objects like strings that we can then translate back into the domain of concepts.

With that in mind, consider a standard word processor, which has only the keys for each alphanumeric character. We can represent any possible text just by typing that text into the program, because texts can be expressed as compositions of individual characters. But notice that the program only actively represents (i.e.\ displays on screen, perhaps requiring some scrolling) a specific string, say the text of \emph{The Hobbit}, when \emph{all of the information content} of \emph{The Hobbit} is typed into it manually\@. \emph{The Hobbit} has about $489,000$ alphanumeric characters, and accounting for some statistical regularities in English \parencite{shannon1951prediction,ren2019entropy}, we can estimate about $587,000$ bits are required to specify its text. Syntactically, that is approximately the amount of information that JRR Tolkien had to create in order to write the book in the first place, and also the amount of information that you receive as new when you read the book. This program would know \emph{essentially nothing} about \emph{The Hobbit} on its own: in order to represent the text, all the characters would have to be received from the environment (i.e.\ the typist's fingers) in the right order.\footnote{Pedantically, even such a word processor knows a small amount about \emph{The Hobbit} because it knows the character set, that text is a string of characters, how to perform spell-checking, etc.} It therefore seems odd to say that the system definitely possesses this text---even though it is potentially available---because essentially all of the information required to select it and represent it comes from outside the program itself. It seems equally odd to say that the system definitely does not possess this text; the equivalent would be to say that a mind definitely does not possess any concept it is not curently representing. It instead seems most plausible to say that \emph{The Hobbit} is at best weakly possessed. It could be represented, but nearly all of the information leading to its representation must come from somewhere else.

Next, consider a program which includes a single button you can click to populate the screen with the entire text of \emph{The Hobbit}. For such a program, very little information is required to come to represent \emph{The Hobbit}: only a decision to click the button. One would, correctly, say that this program essentially possesses the entire text, since it can be triggered with just one click. In fact, if we looked at the source code, we would \emph{have} to be able to discover the text encoded somehow. It might be encrypted or in an unusual format, but the information simply \emph{must be present}, since otherwise the click could not generate the right text. A related result in information theory---the information-processing inequality \parencite{cover1999elements}---implies, roughly, that whatever computation happens when you click the button cannot \emph{create} new information, only reconfigure or delete it.\footnote{Though note that such results are stated, formally, in a probabilistic setting, where information is quantified relative to a distribution of events.} It would therefore be fair to say that this program almost entirely possesses \emph{The Hobbit} for most intuitive senses of possession. The only information missing is the decision to represent \emph{this} text as opposed to the many others that could be typed in character-by-character.

Now, imagine a program with nineteen unlabeled buttons, one for each chapter of \emph{The Hobbit}. Pressing each button appends the text for the corresponding chapter onto the string, but it is up to the user to do so in the right order. Clearly this program encodes \emph{less} information about \emph{The Hobbit} than the previous one, because the user must supply information about chapter order (unlike above, where one click puts them in the correct order). However, as above, the program will necessarily have text of the chapters embedded somehow in its source code. In fact, we can quantify the amount of information required from the user to represent \emph{The Hobbit}, i.e.\ the information needed to choose the buttons in the correct order. There are $19! = 121,645,100,408,832,000$ possible ways in which a user could click each chapter button once, so that selecting the right order (Chapter 1, Chapter 2, Chapter 3, etc.) requires $\log 19! \approx 57$ bits of information to be provided externally (For simplicity, we are neglecting the information required to choose these buttons over e.g.\ keyboard presses.). This program already contains a huge amount of information about \emph{The Hobbit}, but lacks these $57$ bits. To represent the correct text, this information must come from the environment.

Finally, another program might contain a single (unlabeled) button for each of \emph{The Hobbit}'s approximately $4,850$ sentences. This version---you may see where this is going---would require $\log 4,850! \approx 36,316$ bits of information from the user to click the buttons in the right order, meaning that it encodes substantially \emph{less} information about the text than the previous two versions. Even so, it would be easier to click those buttons in the right order than it would be to generate the entire story from scratch using just the keyboard.

It is worth emphasizing that in each of these programs, the compositional closure (i.e.\ the theoretical expressive power) is the same: one can always type any text into the program using the keyboard. The versions differ dramatically, however, in how strongly they possess---both intuitively and by the definition we introduce below---the text of \emph{The Hobbit}, ranging from completely absent to somehow being fully encoded by the source code. What differs between them is how much information is required from the environment in order to represent the text. The required information is essentially an address or an index, picking out one text from the (infinite) space of possibilities. Just as different street addresses pick out different locations, different sets of information lead a learner to select different representations. Moreover, just as a room number is a simpler address than a mailing address, it makes sense that the information required to select a representation depends on the number of possibilities and how they are organized.

The examples above all deal with strings rather than dealing directly with concepts. We are not claiming that to have the concept \textsc{The Hobbit} requires memorizing the entire text of \emph{The Hobbit}. It is entirely plausible that someone who knows nothing of the plot or characters, much less of the book's individual sentences, can nonetheless have some concept, \textsc{The Hobbit}. We use strings because both strings and concepts are computational objects and thus contain a certain amount of information. What transfers is that selecting one concept over others requires information, just as selecting one string over others requires information. Moreover, the amount of information depends not just on the information content of the object being selected or the number of possible objects which could be selected, but the structure of the process by which they are selected. These points are entirely agnostic with respect to theories of conceptual content. Whatever content concepts do in fact have, selecting one concept rather than another requires information. Moreover, the amount of information required may vary considerably based on the internal structure of the hypothesis space, just as the difficulty of producing the text of \emph{The Hobbit} depends on the organization of the word processor.

The amount of information that must come from the environment\footnote{We are using ``environment'' in a fairly broad sense here to mean any source of information outside the specific mechanism used to select and represent concepts. Often, this information might originate outside the learner, as in perception. Other times, it might come from some other part of the learner's mind, as in memory or imagination.} is in fact key to quantifying concept possession. If representing a concept requires very little information from the environment, it is mostly possessed; if it requires a lot of information, it is less possessed.
We can make this idea more precise by first recognizing that a concept $c$ requires some \emph{total} amount of information to specify. That is, whatever theory of conceptual content turns out to be correct, each concept will contain some amount of information, namely the information captured by its computational structure. Even if, as Fodor argued, most lexical concepts are essentially atomic, they must have some computational content such that we can distinguish the atom \textsc{cow} from the atom \textsc{justice}. We are not talking here about information \emph{about} a concept, such as how many stomachs cows have or how justice relates to equity; we are instead making a different kind of claim. If a concept is a computational object, then no matter what theory of conceptual content turns out to be true, and no matter what conceptual structure concepts do in fact turn out to have, they must have some computational structure. That is, they must be some sort of computation. The information required to describe the computation that \emph{is} concept $c$, is $c$'s intrinsic information. Such a measure of intrinsic information, here denoted $T(c)$, is often considered in computer science as Kolmogorov complexity \parencite[e.g.\ the length of the shortest program that will print out $c$;][]{li2008introduction}, formalized in probabilistic inference as surprisal \parencite[e.g.\ the log probability of $c$ in a probabilistic model;][]{shannon1948mathematical}, or characterized in coding theory as description length \parencite[e.g.\ the length of the shortest code for $c$ relative to a coding model;][]{grunwald2007minimum}.\footnote{While here we assume $T(c)$ is fixed, we note that choices in formalizations may affect the precise value of $T(c)$, similar to arguments that the basis of concepts determines what is ``simple'' \parencite{goodman1955fact,piantadosi2016logical}. However, in Kolmogorov (and related) theories, there are results that two different measures of $T(c)$ can differ by at most a constant \parencite{li2008introduction}. While the exact choice of representation language matters in practice---i.e.\ it must respect the mind's finite computational resources---the amount of information in concept $c$, i.e.\ $T(c)$, is still a relatively objective measure.} The details of these definitions, which are largely interchangeable, do not matter for us. What matters is only that there is some sense in which we can say that a concept $c$ requires some total amount of information to specify precisely, independent of a particular learner. For the string, \emph{The Hobbit}, we might use $T(\textit{The Hobbit})=587,000$ from the character-level encoding as a measure of how much information there is, in total, in the text. By contrast, the total information in the concept \textsc{The Hobbit}, i.e.\ $T(\textsc{The Hobbit})$ would almost certainly be some different value.

Second, for a learner, let $R(c)$ be the number of bits of information that the learner must extract from the environment in order to select and thereby represent $c$. These $R(c)$ values were what we computed above for representing the text of \emph{The Hobbit} with each word processor. In general, this could be measured for any learner by quantifying how much information they required, perhaps drawing on the number of examples they needed and the information provided by each example. Note that $R(c)$ is not a measure of how much information a learner must passively encounter but instead of how much they must extract and internalize. Someone might thus need to encounter the same information many times to fully extract the necessary $R(c)$ bits individuating $c$ from related concepts. Moreover, it may be possible that individuals acquire or forget information which restructures their hypothesis space in ways that could raise or lower $R(c)$. Finally, because learners can typically select many concepts to apply in a given situation, $R(c)$ is rarely $0$, even when a concept is familiar, because some information is needed to select one concept from among many.

Then, we can measure the degree of concept possession as the difference between $T(c)$ and $R(c)$:
\begin{equation} \label{eq:possession}
    \text{Degree of possession of } c
    :=
    T(c) - R(c).
\end{equation}
For example, in the word processors above, we might take $T(c)=587,000$ bits to approximately measure how much information there is in \emph{The Hobbit}. The sentence-level word processor would then, at the time of our analysis, have all but $R(c)=36,316$ bits, meaning that program would possess $T(c)-R(c)=550,684$ bits, or $94\%$ of the information required to fully represent the text.

One point worth clarifying here is that the measure of concept possession we propose here may change each time a concept comes to be represented. For example, $R(c)$ might be high the first time someone selects and represents $c$, simply because their concept selection mechanism is not structured to make $c$ easy to represent. Various learning mechanisms---e.g.\ associative learning or Bayesian inference---might restructure the selection mechanism over time. If so, then if $c$ is a commonly selected concept, the information required to select and represent it (i.e.\ $R(c)$) might go down over time. As a result, $c$ would have different degrees of possession at different times and could come to be more strongly possessed as the result of learning.

Note that this measure also allows concepts to vary in their \emph{overall} complexity. This helps us avoid an oddity of binary approaches to concept possession, namely that every possessed concept is as possessed as every other possessed concept. For example, thinking in terms of strings again, this oddity would imply that the text of \emph{The Hobbit} is equally innate to a standard word processor as a much shorter work like \emph{The Raven} or one of Bash\=o's haiku, even though these latter take much less information to specify. Extending this line of thinking, \emph{The Hobbit} would be as innate as one of its sentences. Both of these unworkable facts highlight the futility of trying to reconcile binary senses of innateness with the information required to specify computational structure.

Beyond this, our usage of the term supports a stronger mathematical statement. If $T(c)$ bits are required to select and represent a concept, but only $R(c)$ bits are currently available from the environment, then $T(c)-R(c)$ bits \emph{must} come from the agent. This is why it makes sense to equate $T(c)-R(c)$ with the degree of ``concept possession''. In other words, if you need $T(c)$ total bits but the environment only gave you $R(c)$, then the remaining $T(c)-R(c)$ bits must have come from somewhere---and the agent is the only option.  Similarly, if you harvest $4$ avocados from the environment, and now have $10$ total, then there is no getting around the fact that you had $6$ when you started. There is also no getting around the fact that defining innateness in terms of potential availability, as some have taken Fodor to do, would suggest you always ``possessed'' $10$ (or more) since you always had the potential to have so many! A version of the argument that the information must be present can be made precise relative to probabilistic models and average amounts of information, but we will not undertake that here.

Instead, we consider two more naturalistic examples, starting with indigo bunting navigation. There, the space of hypotheses is relatively constrained because the learning mechanism itself specifies a great deal of information about the form and content of hypotheses. Moreover, only a relatively small amount of information---namely a few observations of the night sky---are needed to fix the location of the north star. Because the information required to identify the north star is likely small compared to the total information contained in indigo bunting navigation, it seems reasonable to conclude that much of the representation is possessed prior to observing the night sky. Avian imprinting, where a small amount of information is used to fix the identity of a caretaker, is arguably similar \parencite{weiskopf2008origins}.

For a workaday concept like \textsc{carburetor}, several cases are worth fleshing out. First, consider a mind currently representing \textsc{carburetor}. In this case, no additional information is required (i.e.\ $R(c) = 0$), because the concept is already represented and so is in that moment fully possessed. Next, consider a mind trying to represent \textsc{carburetor} in terms of a minimal set of Turing-universal primitives. These primitives likely bear little resemblance to \textsc{carburetor}, so selecting the correct hypothesis requires about as much information from environmental input (e.g.\ a class on car repair) as the concept itself contains. In this case, possession will be close to zero, which matches intuition. We can also consider a system which does not contain the content of \textsc{carburetor} in its compositional closure, e.g.\ trying to select the concept using the indigo bunting navigation mechanism. Likely no amount of information will allow this mechanism to form \textsc{carburetor}, so $R(c) = \infty$, and $T(c)-R(c)=-\infty$: the concept \textsc{carburetor} is not only not possessed, it is \emph{anti-possessed}: the current hypothesis space is actively biased against representing it, even when given the correct information. Some human concepts are likely softer versions of this case---for example, a bias to conceptualize events in terms of Newtonian time and space \parencite{spelke2022babies} might make $R(c)$ higher than $T(c)$ for \textsc{relativity} or \textsc{quantum mechanics}. In other words, relativity may be mildly anti-possessed and so take much more information to acquire than to objectively specify (e.g.\ as a program). Similar arguments might apply in cases of conceptual change \parencite{carey1985conceptual,carey2009origins,barner2016core}, where the concepts a learner initially applies to data bias it against a mature conception based on a different set of concepts. Such anti-possession is obvious in retrospect---sometimes it is harder to learn something than it seemingly \emph{should} be---but this possibility cannot be approached easily, if at all, under radical concept nativism.


These examples demonstrate perhaps the most important feature of this proposal: it treats concept possession as a continuous rather than a binary state of affairs. It assesses the degree of possession, including fully possessed, impossible to possess, and everything in between. This variation occurs because both the information required to describe a concept $T(c)$ and the effective amount of information required by a learner $R(c)$ are quantifiable. This gradience connects our notion of concept possession to the empirical phenomena described above, where concepts \emph{do} vary in how hard they are to select or represent. It also meshes nicely with the observation that there are many ways to build in information, including not only explicit representations but also constraints on cognitive architecture and the timecourse of development \parencite{elman1996rethinking}.

Even agreeing with all the analysis so far in this section, one might wonder what relevance it has to the question of concept learning and, more specifically, to the question of radical concept nativism. The key statement we have made is that the degree to which a concept is possessed is a function of how much information must be extracted from the environment in order to represent the concept, relative to its total information content. It seems entirely plausible that this information could be extracted using non-rational processes akin to those hypothesized for triggering. Fodor and others are happy to allow such processes as a way in which experience affects the conceptual repertoire, so one might conclude that our argument is irrelevant to the question of learning.

Such a conclusion, however, would be premature. Bayesian models of learning have come to implement a well-developed and empirically explanatory theory of how information extracted from the environment rationally affects beliefs about a hypothesis space \parencite{tenenbaum2011grow,ullman2020bayesian,lake2017building,griffiths2024bayesian}. In particular, as evidence accumulates, the posterior probability associated with each hypothesis also changes. Some hypotheses become more probable while others become less probable. Moreover, the posterior probability in a Bayesian model directly predicts how much information is required to select that hypothesis from among its competitors \parencite{mackay2003information}. As such, our arguments here show that the kinds of inferences which affect the degrees of belief in a Bayesian posterior also directly affect the degree to which a concept is possessed. In a sense, we are arguing that the arguments separating concept learning from belief fixation are not as strong as they first appeared. Because the degree of possession is graded and context-dependent, any change to the hypothesis space which influences the probability of expressing a concept is a change in the degree to which that concept is possessed. Because Bayesian inference is perhaps the most principled model we have of inductive reasoning, there is every reason to believe that the kind of information extraction we are positing here can be explained as a learning process rather than a sort of sub-psychological brute-causal triggering.

This section introduced a formal and unambiguous proposal for what it could mean to possess a concept. As in earlier sections, our approach has been less to refute a well-known premise in arguments for radical concept nativism and more to argue for notions of innateness and concept possession---on which those premises rely---that better reflect the computational realities of the human mind. In the case of innateness, our solution has been to largely remove it from the discussion. The term remains too ambiguous, both in Fodor's writing and in the field at large, to play a central role. Perhaps more importantly, removing this term helps to shift the focus back onto the question of learning and, in particular, to the information we use to represent our concepts. The notion of concept possession presented here captures a key empirical fact about human cognition, namely that people's cognition is sensitive to the information they extract from their environment. The result is a view of concept possession which is incompatible with radical concept nativism. In particular, if the degree to which a concept is possessed depends on the amount of information currently possessed relative to the concept's total information content, then it becomes possible to learn concepts. Any learning episode which changes the amount of information required to represent a concept becomes a form of concept learning.

\section{Discussion} \label{sec:discuss}

Our goal in this paper has been to work out some of the implications of a computational theory of human psychology, specifically the implications for concept learning and radical concept nativism. Our approach has not been to dismiss radical concept nativism simply by defining the problem away. It has instead been to take the motivating concerns seriously enough to note places where existing arguments fail to take into account key aspects of what human cognition must be like. To close these gaps, we then proposed alternative treatments which are better aligned with current evidence and which, together, present a strong challenge to the arguments for radical concept nativism.

In particular, we made three interrelated proposals involving expressive power, conceptual structure, and concept possession. For expressive power, we argued that because humans are Turing-universal, they have maximal expressive power, making moot any question of increasing expressive power through learning. For conceptual structure, we observed that if concepts are computational objects, then they can be described as having complex internal structures. These structures are not necessarily best described in terms of simpler concepts (though perhaps in some cases, that might make sense) but can certainly be described in terms of simpler computations, such as the neural activity in terms of which biological minds are implemented. For concept possession, we proposed that concept possession is best measured in terms of the information required to select or represent a concept relative to that concept's total information content.

Taken together, these three proposals paint a different picture of cognitive development than the one assumed in arguments for radical concept nativism. On this view, human minds are maximally expressive computational systems. This expressiveness helps to explain both the rich variety of structures people learn \parencite{turing1936computable,chater2013programs,piantadosi2016four,dehaene2022symbols} and their behavior during learning \parencite{goodman2008rational,lake2015human,rule2020child,planton2021theory}. The expressiveness comes at least in part from the use of a set of Turing-universal computational primitives (the set can be quite small, though need not be). To reiterate, the claim here is not that people are born with some small set of concepts in terms of which all others are defined. We agree that it seems unlikely that the search for such a set will be successful. Instead, the arguments here demonstrate that it is possible to define all concepts in terms of simpler, subconceptual computations. Such resources would not themselves be concepts but would provide the building blocks from which concepts could be constructed. While the actual primitive basis of conceptual thought is an open empirical question, it seems at least plausible to us that the basis is not a enormous set of built-in atomic concepts but instead a much smaller set of subconceptual computations.\footnote{Note, too, that we are not claiming that humans only have access to semantically simple, subconceptual primitives. Our proposal is entirely compatible with the existence of semantically complex core cognitive resources for reasoning about objects, agents, quantity, and geometry \parencite{spelke2007core,barner2016core}. Subconceptual primitives could in fact provide the machinery necessary to explain how core cognitive representations could be incorporated into full-blown concepts.} These computational primitives may be so simple that it's difficult to imagine how they could be used to express concepts, beliefs, etc. \parencite[see, e.g.][]{piantadosi2021computational}. But, as is the case with the modern digital computer, complex compositions of simple parts can be used to express a great diversity of abstractions. Viewing the mind computationally, it makes sense to at least give serious consideration to psychological theories built from extremely simple computational parts.

If human minds have access to Turing-universal primitives, then development hinges not on acquiring new expressive power but on learning to manage the expressive power that is already available. In this case, conceptual development largely becomes a matter of restructuring the space of possible concepts to provide easy access to the concepts which best explain the world a mind observes and which help it leverage that world in pursuit of its own ends. That is, a great deal of developmental effort is spent searching through the computational landscape of all the things a mind could think for those few things which are most useful to think given what that mind has seen. Cognitive development thus largely becomes a process of exploring and restructuring an unboundedly large and wildly complex space of computations.

Part of the claim here is that any concept can be expressed, provided sufficient information to select it from the sea of other possible concepts. The business of cognitive development is then one of adjusting the amount of information required for selection so that relatively little is needed for those things which are likely to be most useful and relatively more is required for the rest. Over time, a few bits extracted from the environment become sufficient to cue relevant concepts, even if those concepts contain far more than just a few bits. Seen this way, the study of concept learning is not about drawing binary distinctions separating the concepts we possess from those which we do not possess. It is instead about tracking subtle shifts in the degree to which we possess certain concepts; that is, subtle shifts in the ability for a few bits of environmental information to lead to the rational representation of much more complex structures.

Existing discussions of radical concept nativism focus on how individual concepts could go from being categorically not possessed to categorically possessed. Our view suggests that concept possession is about the amount of information required to represent a concept, meaning that concept possession is not categorical at all. It is instead continuous. Any change which decreases the information required to represent a concept is part of concept acquisition. These changes need not be as dramatic as expanding expressive power by adding new primitives: any decrease in the information required to represent a concept counts as partial acquisition, making the line between concept learning and other sorts of learning much fuzzier than Fodor assumed. Even the learning Fodor discounts as mere ``belief fixation'' can be seen as concept acquisition, because changes in the degree of belief assigned to a hypothesis affect the information required to select it from other hypotheses \parencite{mackay2003information}. As such, the characterization of learning mechanisms of all kinds becomes central to the study of concept learning. More importantly, when concept possession is understood in terms of the information required to represent a concept, concept learning becomes commonplace.

Despite the mind's incredibly expressive primitives, it is still subject to strict resource limits in terms of the memory and computation which individuals can devote to storing, expressing, and activating cognitive structures. While Fodor's arguments explicitly reject resource limits as a central feature of the debate about concept learning, doing so was a mistake. It is essential to understand that, in reality, processes like chunking, abstraction, and library learning are important tools for exceeding resource limits. These processes also make it easier to refer to some parts of the space (i.e.\ learned chunks) than others (i.e.\ those remaining outside resource bounds), performing exactly the kinds of restructuring this view places at the heart of learning and development. In practice, these processes are already enough to provide for strong forms of concept learning in that minds can use them to express concepts which were not previously expressible given strong resource limits.

One other key shift in perspective here is that if concepts are built from simpler, subconceptual parts, they share much in common with other kinds of mental representations. While they perhaps play a different role in cognition than beliefs or behavioral schemas, the developing mind faces the same problem for each: picking out particular structures from the space of all possible computations which best explain its observations and serves its goals. If so, then many of the values and techniques which serve learners in one area, e.g.\ developing a theory of social norms, may also be helpful in other areas, e.g.\ learning concepts of natural number \parencite{rule2020child}.

Much of Fodor's work on nativism, and on concepts more generally, appears to be concerned with fundamentally empirical issues. For example, he sees the fact that people can flexibly compose concepts as an important criterion that quickly rules out many of the proposed theories of concepts. He also sees the empirical difficulties with defining concepts as a key problem that must be explained. We close this section by considering two ways in which our proposal speaks to these concerns.

First, as discussed above, composition is a basic feature of many Turing-universal formalisms. Structures in such systems---particularly functional frameworks such as lambda calculus or combinatory logic---are designed to compose with one another. To build some intuition, consider the case of mathematical functions, $f(x) = x^{2}$, $g(x) = x-1$, $h(f, x) = f(f(x))$, $j(x) = h(f,x)$, and $k(x) = h(g,x)$. Although $j$ and $k$ have similar structure, they compute different functions. The reason for the difference has to do with the internal structure of $f$ and $g$. Because $f$ and $g$ specify different internal computations, composing them in different ways leads straightforwardly to different behaviors. Similarly, if concepts and other high-level mental constructs are also composed from simpler computations, it would be not at all surprising to learn that they compose in complex and powerful ways \parencite{piantadosi2021computational}. Consider, again the earlier example of \textsc{blue sky} and \textsc{sky blue}. One refers to any sky which is any shade of blue, while the other refers to a particular shade of blue (i.e.\ the color of some ideal sky). The idea here is that the complex computational structure of the concepts allows \textsc{blue} to act on \textsc{sky} in one way while \textsc{sky} applied to \textsc{blue} acts in yet another way. Some of these effects may be particular to specific compositions, while others may apply much more broadly because of the underlying computational structure.

Second, the idea of characterizing concepts in terms of subconceptual parts also helps to make sense of Fodor's discussion of definitions. In short, concepts have definitions (i.e.\ the composition of subconceptual parts associated with them), but we may not be able to express or articulate them in terms of the higher-level concepts to which we have verbal access. Instead, describing an object built from subconceptual parts entirely in terms of conceptual parts may be a complex and error-prone computation where important nuance and structure are necessarily lost in translation. That is, \textsc{to paint} may be so hard to define in English because the concept is not defined in terms of \textsc{paint}, \textsc{brush}, or \textsc{surface} but in terms of computations which we have yet to name or conceptualize \parencite{piantadosi2024concepts}.

\section{Conclusion}

While arguments for radical concept nativism have spurred many of the field's greatest minds to valuable insights, they serve as something of a cautionary tale. Dozens of papers across fifty years consider arguments whose logic ultimately does not reflect the situation in which human learners find themselves. Our approach in this paper has been to first articulate key places in which the learning they consider differs from human learning, and then to provide coherent alternatives.

The cumulative effect of our arguments has been to demonstrate that human cognitive development takes place in a very different context from the one which radical concept nativism supposes. Formal tools like information measures and models of computation allow us to create clear and unambiguous accounts of how both learning and preexisting content combine. They show that highly expressive learners draw on information from both internal and environmental sources to learn new representations with complex conceptual content.

\printbibliography

\end{document}